\documentclass{article}

\usepackage{PRIMEarxiv}

\usepackage[utf8]{inputenc} 
\usepackage[T1]{fontenc}    
\usepackage{hyperref}       
\usepackage{url}            
\usepackage{booktabs}       
\usepackage{amsfonts}       
\usepackage{amsmath}       
\usepackage{nicefrac}       
\usepackage{microtype}      
\usepackage{lipsum}
\usepackage{fancyhdr}       
\usepackage{graphicx}       
\graphicspath{{media/}}     
\usepackage{listings}
\usepackage{cleveref}
\usepackage[table,xcdraw,dvipsnames]{xcolor}
\usepackage[inline]{enumitem}
\usepackage{adjustbox}
\usepackage{pifont}
\usepackage[para]{threeparttable}
\usepackage[inline]{enumitem}
\usepackage{multirow}
\usepackage[normalem]{ulem}
\usepackage{subcaption}
\usepackage[tableposition=bottom, figureposition=top]{caption} 
\usepackage{fancyvrb}
\usepackage{tikz}

\definecolor{codebg}{RGB}{250,250,250}
\definecolor{commentcolor}{RGB}{106,153,85}
\definecolor{keywordcolor}{RGB}{0,0,180}
\definecolor{stringcolor}{RGB}{170,0,0}
\definecolor{numbercolor}{RGB}{128,0,128}

\definecolor{pastel1}{HTML}{ADD8E6}
\definecolor{pastel2}{HTML}{F2A65A}
\definecolor{pastel3}{HTML}{6F8F72}
\definecolor{pastel4}{HTML}{D2C4B4}

\lstdefinestyle{mystyle}{
    backgroundcolor=\color{codebg},
    commentstyle=\color{commentcolor}\ttfamily,
    keywordstyle=\color{keywordcolor}\bfseries,
    numberstyle=\tiny\color{gray},
    stringstyle=\color{stringcolor},
    basicstyle=\ttfamily\footnotesize,
    breakatwhitespace=false,
    breaklines=true,
    captionpos=b,
    keepspaces=true,
    numbers=left,
    numbersep=8pt,
    showspaces=false,
    showstringspaces=false,
    showtabs=false,
    tabsize=4,
    frame=single,
    rulecolor=\color{gray},
    moredelim=**[is][\color{numbercolor}]{@}{@},
    escapeinside={|}{|},
}
\useunder{\uline}{\ul}{}

\newcommand{\repourl}{https://github.com/pablomiralles22/paper-nli-grpo}
\newcommand{\hf}[1]{\href{https://huggingface.co/#1}{\emph{#1}}}
\newcommand{\hfd}[1]{\href{https://huggingface.co/datasets/#1}{\emph{#1}}}

\newcommand{\databar}[5]{
  \begin{tikzpicture}[baseline={([yshift=-.5ex]current bounding box.center)}]
    \fill[#4!40] (0,0) rectangle ({#1/#2*#3},0.5);
    \draw (0,0) rectangle (#3,0.5);
    \node at (#3/2,0.25) {#5}; 
  \end{tikzpicture}
}

\usepackage[
    style=numeric,
    sorting=none,
    maxbibnames=10,
    giveninits=true,
    natbib=true
]{biblatex}
\title{Evaluating the Scalability and Adversarial Generalization of GRPO-Trained NLI Models}

\author{
  \textbf{Pablo Miralles-González\textsuperscript{1}},
  \textbf{Javier Huertas-Tato\textsuperscript{1}},
  \textbf{Alejandro Martín\textsuperscript{1}},
  \textbf{David Camacho\textsuperscript{2}}
\\
\\
  \textsuperscript{1}Department of Computer Systems,
  Technical University of Madrid,
  Madrid, Spain
\\
  \textsuperscript{2}United Arab Emirates University,
  Al Ain, United Arab Emirates
}

\begin{document}
\maketitle

\begin{abstract}
Natural Language Inference (NLI) is a central task in natural language understanding with applications in fact-checking, question answering, and information retrieval. Despite its importance, current NLI systems heavily rely on supervised learning with datasets that often contain annotation artifacts and biases, limiting generalization and real-world applicability. In this work, we apply a reinforcement learning-based approach using Group Relative Policy Optimization (GRPO) for Chain-of-Thought (CoT) learning in NLI, eliminating the need for human-labeled rationales and enabling this type of training on challenging datasets such as ANLI. We fine-tune 7B, 14B, and 32B language models using parameter-efficient techniques (LoRA and QLoRA), demonstrating strong performance across standard and adversarial NLI benchmarks. At the 32B scale, GRPO-trained models generalize better than other supervised baselines in adversarial sets. With AWQ quantization, the 32B model fits within 22GB of CUDA memory. This work provides a scalable and practical framework for building robust NLI systems without sacrificing inference quality.
\end{abstract}
\keywords{Natural Language Inference \and LLMs \and GRPO}

\section{Introduction}\label{sec:intro}

Natural Language Inference (NLI) is a core task in natural language understanding, where the goal is to determine whether a given premise entails, contradicts, or is neutral with respect to a hypothesis. NLI has critical applications in real-world settings such as automated fact-checking, question answering, and information retrieval, where accurate reasoning over textual content is essential.

Despite its importance, progress in NLI has been mostly reliant on supervised learning approaches, which require large amounts of labeled data. However, existing NLI datasets are limited in both size and quality. They often exhibit annotation artifacts and spurious patterns, such as the hypothesis-only bias, that models can exploit without true reasoning. As a result, these datasets do not provide a reliable foundation for learning robust inference capabilities.

Although much research has focused on improving the quality of the data or mitigating biases during training, only marginal gains have been achieved and real-world applicability is still far away. We believe that if base transformer models do not acquire sufficient semantic understanding from pretraining, they will not be able to learn the task through fine-tuning on small, biased datasets. The complexity of the task exceeds what can be learned from such limited supervision.

In this work, we apply modern reinforcement learning techniques, specifically Group Relative Policy Optimization (GRPO), to Chain-of-Thought (CoT) learning for Natural Language Inference using larger language models (7B, 14B, and 32B parameters) with parameter-efficient fine-tuning via LoRA and QLoRA. Unlike prior approaches that depend on labeled rationales. such as those built on e-SNLI, this method requires no explanation annotations, enabling application to more challenging datasets like ANLI.

Our results show that even relatively small base models (7B) already generalize better to adversarial NLI benchmarks than prior smaller supervised models, which often overfit to artifacts. GRPO training leads to consistent gains across both standard (SNLI, MultiNLI) and adversarial benchmarks (ANLI, Counter NLI, HANS, NLI Diagnostic), particularly in quantized settings where LoRA reduces performance loss due to compression. 



We also compare GRPO to SFT without CoT, supervised cross-encoder classification and generic reasoning models. Our 32B AWQ-quantized GRPO-trained model reaches the best results across adversarial NLI tasks in our experiments, while using 22GB of CUDA memory. These results indicate that quantized LLMs fine-tuned with GRPO and LoRA represent a practical path toward deploying robust NLI systems in memory-constrained environments, without sacrificing inference quality.

Our main contributions are the following.
\begin{itemize}
    \item We propose a novel application of Group Relative Policy Optimization (GRPO) for Chain-of-Thought (CoT) learning in Natural Language Inference (NLI), removing the need for labeled rationales. This approach enables training on challenging datasets like ANLI without requiring explanation annotations, broadening applicability beyond rationale-augmented datasets such as e-SNLI.

    \item We train 7B, 14B and 32B AWQ-quantized models with QLoRA using our methodology, and provide all adapter weights publicly. Our 32B AWQ-quantized model achieves great results on adversarial NLI tasks within 22GB of CUDA memory.

    \item We compare GRPO to SFT without CoT, supervised cross-encoder classification and generic reasoning models for NLI tasks.

    \item We thoroughly study and analyze the impact of our training methodology across model sizes, number of trainable LoRA parameters and weight precisions. Our results highlight the importance of using stronger and larger base models if we want true generalization, and show that GRPO training provides strong and robust performance gains in NLI across standard and adversarial datasets.

    \item We show and analyze the training dynamics we found during GRPO training, as well as the differences found in model outputs after training our 32B model.

\end{itemize}

The code is publicly available at \url{\repourl}.

\section{Background}\label{sec:background}

Natural Language Inference (NLI) is a core task in natural language understanding that involves determining the logical relationship between a premise and a hypothesis---typically classified as entailment, contradiction, or neutral. Despite its importance for very relevant and current problems such as automated fact-checking and its versatility for zero-shot predictions in many tasks, the field of Natural Language Inference has not seen many advances in the last few years, likely overshadowed by a focus on LLM reasoning on complex tasks like mathematics or coding.

Most work focuses on supervised fine-tuning of Language Models. However, the available labeled data is scarce and low in quality, and the task is very complex. Although researchers have tried to mitigate this problem in several ways, these approaches yield only marginal performance improvements.
We also find approaches that apply decoder-only LMs to NLI, but current research efforts around these models focus on more complex reasoning tasks.

In this section, we go over the most relevant work on Natural Language Inference, and also briefly cover the most relevant reasoning research for our work.

\subsection{Natural Language Inference datasets}

We begin our discussion by exploring the most relevant data sources for NLI. These datasets have been instrumental in training and evaluating models, but they are also a source of limitations in how current models learn and generalize.

\begin{itemize}
\item \textbf{SNLI}~\citep{bowman2015largeannotated} was the first large-scale NLI dataset, with 570K human-written and manually annotated text pairs. As most labeled datasets in this section, each pair includes a premise and a hypothesis, labeled with one of the three NLI categories: entailment, neutral or contradiction.

\item \textbf{e-SNLI}~\citep{camburu2018eSNLINatural} extends SNLI by adding natural language explanations to each annotated pair. These explanations were crowd-sourced and provide a rationale for the label assigned to each example. The inclusion of explanatory text introduces a richer and more interpretable training signal, allowing models not only to predict labels but also to justify their predictions.

\item \textbf{MultiNLI}~\citep{williams2018BroadCoverageChallenge} improves SNLI by offering a broader range of linguistic variation. It contains 433,000 sentence pairs drawn from ten different genres of written and spoken English, such as fiction, telephone speech, and government reports. This genre diversity aims to test the generalization ability of NLI models across more realistic and heterogeneous text sources.
\end{itemize}

However, despite their scale and utility, these early NLI datasets were found to contain significant annotation artifacts and biases. Models were able to achieve surprisingly high accuracy even when trained on hypotheses alone, without access to the premise~\citep{tsuchiya2018Performanceimpact,gururangan2018AnnotationArtifacts,poliak2018HypothesisOnly}. These findings suggest that superficial cues in the hypothesis---rather than genuine inferential reasoning---were often sufficient for correct classification, thereby undermining the intended goal of evaluating deep language understanding.

To probe deeper aspects of model understanding and mitigate shallow learning patterns, several diagnostic and adversarial datasets have been proposed. These aim to evaluate whether NLI models rely on genuine reasoning or exploit superficial cues in the data.

\begin{itemize}
\item \textbf{NLI Diagnostic}~\citep{wang2018GLUEMultiTask} is a small but carefully constructed set of 550 hand-crafted sentence pairs included in the GLUE benchmark. Unlike large-scale datasets, this collection is designed to stress-test specific linguistic phenomena such as commonsense and world knowledge, logical inference, predicate-argument structure, and lexical semantics. The examples are drawn from multiple domains to enhance coverage and reduce dataset-specific biases.

\item \citet{mccoy2019RightWrong} demonstrated that models trained on MultiNLI tend to rely on simple syntactic heuristics---such as lexical overlap, subsequence matching, and constituent alignment—to make predictions. To expose this issue, they created the \textbf{HANS} dataset by generating sentence pairs from templates that systematically exploit and challenge these heuristics. Models that performed well on MultiNLI demonstrated poor performance on HANS, especially on non-entailment cases, revealing a lack of robust generalization.

\item \textbf{Counter-NLI}~\citep{Kaushik2020Learning} addresses the hypothesis-only bias in SNLI by introducing counterfactual reasoning. Each example is augmented with two human-written, validated counterfactuals: one modifying the premise and the other the hypothesis. This structure encourages models to pay attention to both sentences in the pair and reduces over-reliance on spurious lexical cues. Models trained with this data have shown improved robustness to this common NLI pitfall.

\item \textbf{ANLI}~\citep{nie2020AdversarialNLI} was created using the Human-And-Model-in-the-Loop Enabled Training (HAMLET) framework, which rethinks dataset construction by placing humans in an adversarial role. Instead of relying on automatic perturbations, human annotators are asked to craft hypotheses specifically designed to fool a trained NLI model. When successful, they also provide explanations for the model’s errors, offering insight into the model’s weaknesses. This iterative process was repeated in three rounds, each using a stronger model and a broader domain of premises. The result is a multi-part dataset of increasing difficulty, effectively exposing the limitations of current systems and promoting more robust learning.

\item \citet{liu2022WANLIWorker} introduced a complementary approach with \textbf{WANLI}, blending large language models and human oversight. Using dataset cartography, the authors first identified challenging examples in MultiNLI to guide generation. GPT-3 then produced similar sentence pairs, which were filtered, edited, and labeled by human annotators. The final dataset consists of 107,885 high-quality examples that cover a wide range of reasoning patterns. Despite being smaller than MultiNLI, WANLI led to superior generalization: models trained on it outperformed MultiNLI-trained baselines on eight out-of-domain benchmarks, achieving notable gains—11\% on HANS and 9\% on ANLI.
\end{itemize}


\subsection{Supervised methods for Natural Language Inference}

In this section, we focus on supervised learning methods. As in many NLP tasks, state-of-the-art solutions rely on pre-trained transformer models~\citep{vaswani2017AttentionAll}. Given the nature of the NLI task, the most suitable architecture involves applying an encoder to both the premise and hypothesis together. The common practice is to use encoder-only transformers, such as BERT~\citep{devlin2019BERTPretraining}, RoBERTa~\citep{liu2019RoBERTaRobustly}, or DeBERTa~\citep{he2021deberta}. Alternatively, the task can be reformulated in a text-to-text format using encoder-decoder models like T5~\citep{raffel2020Exploringlimits} or T0~\citep{sanh2022multitask}.

However, as noted earlier, these models often perform poorly when fine-tuned on small, biased datasets---especially for complex tasks with high model capacity. This has led to increased research interest in improving data quality and refining training procedures to enhance generalization across different data distributions. Unfortunately, these approaches lead to marginal improvements, but generalization to out-of-domain and adversarial examples is still very poor.

\subsubsection{Using synthetic data}
\citet{he2024UsingNatural} introduced GNLI, a dataset of 671K NLI examples spanning 38 domains, generated using large language models (LLMs) and their text generation capabilities. They observed some gains in the TRUE benchmark\citep{honovich2022TRUEReevaluating}, which evaluates factual consistency. To assess GNLI’s effectiveness for NLI, they trained and evaluated T5 models of various sizes on different combinations of MultiNLI, ANLI, WANLI, and GNLI. Results show that GNLI occasionally improves cross-dataset generalization, but overall gains are modest. Notably, using GNLI alongside the original data slightly improves test performance in larger models (3B and 11B), which aligns with expectations.

\subsubsection{Dataset debiasing}
\citet{liu-etal-2020-hyponli} identified artificial patterns in hypothesis sentences that correlate spuriously with specific labels. They removed instances containing these patterns from the training data through downsampling. This led to improved performance on challenging SNLI and MultiNLI examples compared to random downsampling, but overall performance decreased.

\citet{pmlr-v119-bras20a} introduced \textbf{AFLITE}, a greedy debiasing method that iteratively filters out the easiest examples for a trained classifier. Training RoBERTa on the filtered SNLI data yielded substantial gains on the HANS dataset, with smaller improvements on ANLI and the NLI diagnostics subsets focused on logic and lexical semantics.

\citet{wu2022GeneratingData} proposed generating synthetic data using a fine-tuned LLM, then applying \textbf{z-filtering} to reduce dataset bias. This method removes examples to ensure that label distributions are uniform with respect to known biasing features (e.g., unigrams, hypothesis length, or BERT hypothesis-only predictions). Their debiased dataset produced slight improvements on SNLI and MNLI when training BERT-based models.

\subsubsection{Preventing biases during fine-tuning}
\citet{liu2020EmpiricalStudy} compiled a range of known biases learned by NLI models from prior work. For each bias, they trained a dedicated bias-only model, which was then used in model-agnostic debiasing strategies. These included re-weighting training examples based on bias-model predictions and ensembling logits from both the bias-only and target models. They also explored data augmentation via word masking and substitution. Their findings show that while such debiasing methods can slightly mitigate several biases, and no approach mitigates all of them.

\citet{liu-etal-2020-hyponli} also explored adversarial training to reduce hypothesis-only bias. They designed a model that encodes premise and hypothesis separately, with a final classification head over both. A discriminator was added to predict the label from the hypothesis representation alone, and the main model was trained to fool this discriminator while maintaining correct predictions. This adversarial setup, along with other modifications, improved performance on challenging SNLI and MNLI examples but led to a drop in overall accuracy.

\citet{wang2021infobert} introduced two key methods. First, they applied the information bottleneck principle to balance model accuracy and complexity. Second, they proposed the \textbf{Anchored Feature Regularizer}, which uses gradient norms with respect to the input embeddings to identify token importance. Tokens with small gradients are considered less influential, while those with large gradients are more vulnerable to adversarial changes. They then align the model's global representation with that of a ``middle-ground'' token using a mutual information-based objective. These techniques improved model performance on ANLI and other adversarial benchmarks.

\citet{stacey2022supervising} proposed guiding model attention using human-written natural language explanations. They introduced a loss term that encourages the model's final-layer [CLS] attention to focus on tokens relevant to these explanations. This supervision led to better generalization both in- and out-of-distribution. Additionally, they explored an alternative approach by adding an external attention layer over the transformer's output, rather than modifying the internal attention mechanisms.

\citet{kavumba2023Promptingexplanations} examined whether prompting models for explanations enhances performance on adversarial NLI tasks. They trained encoder-decoder Transformer models (T5-3B\citep{raffel2020Exploringlimits} and T0-11B~\citep{sanh2022multitask}) on the e-SNLI~\citep{camburu2018eSNLINatural} dataset to predict both the label and the explanation. Unlike Chain-of-Thought methods, which predict the label after the explanation, their approach requires the model to make the prediction first.
Even though the explanation doesn't influence the prediction, it provides a new and more complex training signal that can lead the model to understand the task properly instead of relying on spurious correlations.
Their results show that this method reduces reliance on spurious cues and significantly improves performance on adversarial datasets, setting new state-of-the-art results in ANLI~\citep{nie2020AdversarialNLI}, Counter-NLI~\citep{Kaushik2020Learning}, and NLI Diagnostic~\citep{wang2018GLUEMultiTask}. However, they have not released their code, and our attempts to replicate their results resulted in much worse performance (see \ref{sec:anli-sota}).

The idea of training with explanations on e-SNLI was previously explored by \citet{narang2020WT5Training}, who trained the WT5-11B model on several text-to-text tasks with explanations, achieving an impressive 92.3\% accuracy on the e-SNLI test set.

\subsection{Non-supervised methods for Natural Language Inference}

We find a minority of work on non-supervised methods. For example, \citet{he2024UsingNatural} introduced a method that leverages machine-generated natural language explanations as in-context exemplars to enhance large language model (LLM) performance on the target instance. By conditioning the LLM on multiple explanation-augmented examples, their approach enables the model to generalize more effectively to unseen, unlabeled inputs. The authors demonstrate consistent performance improvements across both open-source and proprietary LLMs, as well as across several adversarial natural language inference (NLI) benchmarks, highlighting the robustness and generality of their method. Despite this, overall performance still has great room for improvement.

\subsection{Reinforcement learning for reasoning}

There has been a great body of research on improving the reasoning capabilities of LLMs, specially by encouraging the models to generate chains of thought~\citep{wei2022chain} before answering. The use of large-scale training data has been the main driving force behind reasoning improvements, with pre-training datasets like~\citep{wang2024MathPilebilliontokenscale} or curated instruction data~\citep{yue2024mammoth}.

The most relevant for our work is the successful application of large-scale reinforcement learning to improve reasoning capabilities in complex tasks like mathematical or coding problems. The most prominent example is the DeepSeek-R1 model~\citep{deepseek-ai2025DeepSeekR1Incentivizing}, trained with the GRPO~\citep{shao2024DeepSeekMathPushing} algorithm. However, to the best of our knowledge, these new state-of-the-art techniques have not yet been explored in the particular case of NLI.


\section{Method}\label{sec:method}

\begin{figure}
    \centering
    \includegraphics[width=1.0\linewidth]{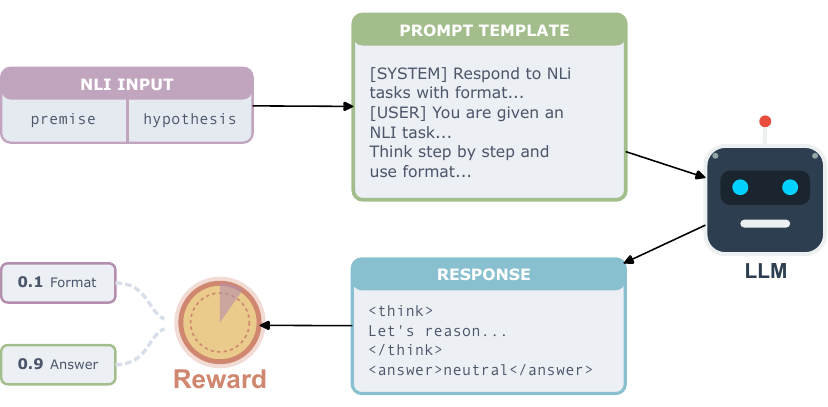}
    \caption{Diagram of our NLI problem set-up, prompting the model to produce a rationale before the final prediction in a structured format. A small reward is given for producing the correct format, and a bigger one for predicting the correct answer.}
    \label{fig:method}
\end{figure}

Similar to prior work, we formulate the NLI task as a text-to-text problem. In particular, we prompt the model to produce an explanation or rationale that conditions the final prediction. The process is shown graphically in \cref{fig:method}. Since we train with reinforcement learning, we do not require manually produced explanations, but use those sampled online from the model. This allows us to use more difficult datasets, like \emph{ANLI}, for training. What is more, our training objective includes a KL-divergence term that prevents the model from diverging too much from its initial state. This reduces the risk of bypassing the explanation and predicting based on the premise-hypothesis pair, mitigating the reliance on spurious correlations. The following subsections explain our training process in more detail.

\subsection{Reinforcement learning with GRPO}

For reinforcement learning we decide to use \emph{Group Relative Policy Optimization} (GRPO)~\cite{shao2024DeepSeekMathPushing} due to its lower memory usage, as it removes the need for a critic model. Given a prompt $p$, GRPO samples a group of completions $\{o_1, o_2, \dots, o_G\}$ from the old policy $\pi_{\theta_{old}}$, with their respective rewards $\{r_1, r_2, \dots, r_G\}$. The GRPO objective is given by:

\begin{multline}
\label{eq:grpo}
\mathcal{L}_{\text{GRPO}}(\theta) =
\frac{1}{G} \sum_{i=1}^G \left[
    \min\left( \frac{\pi_\theta(o_i|p)}{\pi_{\text{old}}(o_i|p)} A_i, \text{clip}\left(\frac{\pi_\theta(o_i|p)}{\pi_{\text{old}}(o_i|p)}, 1 - \epsilon, 1 + \epsilon \right) A_i \right)
\right] \\
- \beta \, \text{KL}(\pi_\theta \| \pi_{\text{ref}})
.\end{multline}
Here:
\begin{itemize}
    \item \( A_i\) is typically the normalized advantage \(\frac{r_i - \mu}{\sigma}\), with \( \mu \) and \( \sigma \) being the mean and standard deviation of rewards \( r_i \) over the group. However, we follow the recommendations in Dr. GRPO~\cite{liu2025UnderstandingR1ZeroLike} and remove the division by $\sigma$. This makes the optimization process give the same weight to all examples, instead of focusing on those with less diversity. Since our rewards are in the range $[0,1]$, we do not expect any stability problems to arise.
    \item \( \epsilon \) is a hyperparameter that controls a clipping range. This clipping prevents the update from becoming too large when the training policy deviates too much from $\pi_{old}$.
    \item \( \beta \) is a regularization coefficient for the KL divergence term between the training policy and the reference policy \( \pi_{\text{ref}} \), in our case the original or base model. This term stops the model from diverging too much from the base model, which reduces the risk of reward hacking.
\end{itemize}

\subsection{Problem formulation}

Given an NLI problem or premise-hypothesis pair, we formulate it as a text-to-text problem with the prompt template shown in \cref{lst:prompt}.
Our approach requires the model to generate an explicit reasoning step before producing a final prediction, enforced through the reward function. This ensures that the final output is tightly conditioned on the explanation, reducing reliance on spurious correlations between the premise and hypothesis. While the model could theoretically ignore the reasoning process or explanation, this is mitigated in two ways: predictions that contradict the explanation are rarely sampled, and the KL divergence loss penalizes deviations from the base model’s output distribution, discouraging such behavior.

\begin{lstlisting}[
    caption={Prompt template used for NLI premise-hypothesis pairs.},
    label={lst:prompt}
]
|{\color{Cyan}[SYSTEM]}| Respond in the following format:
<think>...</think>
<answer>contradiction/entailment/neutral</answer>
|{\color{Cyan}[USER]}| You are given a Natural Language Inference (NLI) task. You are given a premise and a hypothesis, and you need to determine whether the hypothesis is a contradiction, entailment, or neutral given the premise.
Premise: |{\color{Lavender}\{premise\}}|
Hypothesis: |{\color{Lavender}\{hypothesis\}}|
Given the premise, is the hypothesis a contradiction, entailment, or neutral? Think step-by-step.
Use the following format for your response:
<think>...</think>
<answer>contradiction/entailment/neutral</answer>
\end{lstlisting}

We also experimented with more structured prompts involving step-by-step instructions, but these yielded longer completions without performance gains. Our final prompt offers the model greater flexibility, enabling broader exploration of the solution space. Finally, we note that some redundancy is deliberately included in the instructions to ensure that even small, untrained models adhere to the required output format.

\subsection{Reward function for GRPO}

To optimize the policy via Group Relative Policy Optimization (GRPO), we define a multi-objective reward function $R$ consisting of a format reward $r_{f}$ and a correctness reward $r_{c}$:
\begin{equation}
    R(y) = r_{f}(y) + r_{c}(y)
\end{equation}
where $y$ is the model-generated completion. 

\paragraph{Format Reward ($r_{f}$)} 
We incentivize the model to follow a specific Chain-of-Thought (CoT) structure. A reward of $0.1$ is granted if the output strictly conforms to the format: \texttt{<think>...</think> <answer>\{label\}</answer>}.

\paragraph{Correctness Reward ($r_{c}$)} 
The primary objective remains the accuracy of the inference task. The model receives a reward of $0.9$ for providing the correct label relative to the ground truth. 

\subsection{Hyperparameters}\label{ssec:hyperparameters}

In \cref{tab:hyperparams} we show the hyperparameters we used for GRPO training. We did not fit the GRPO hyperparameters to optimize the results due to resource constraints, but used standard values. Hyperparameters like the number of generations per prompt, the batch size or the number of optimization steps per batch were selected to fit the training process in limited hardware.

\begin{table}[!ht]
    \centering
    \caption{GRPO training hyperparameters. Where two values are provided, the second value applies to the 14B and the 32B models, as well as the 7B model in \cref{tab:results-by-model-size}. The first value applies to the rest.}
    \label{tab:hyperparams}
    
    \begin{tabular}{r|cc}
\toprule
Max steps                       & 1000          & 4000         \\
Batch size                      & 32            & 8            \\
\# gens per prompt              & \multicolumn{2}{c}{4}        \\
KL Beta                         & \multicolumn{2}{c}{0.01}     \\
\# optimization steps per batch & \multicolumn{2}{c}{4}        \\
Clipping epsilon                & \multicolumn{2}{c}{0.2}      \\
Learning rate                   & \multicolumn{2}{c}{$5\cdot 10^{-6}$} \\
Max grad norm                   & \multicolumn{2}{c}{1}        \\
Warmup steps                    & \multicolumn{2}{c}{50}       \\
Scale rewards                   & \multicolumn{2}{c}{False}    \\
Training precision              & \multicolumn{2}{c}{BF16}     \\
\bottomrule
\end{tabular}
\end{table}

\subsection{Training data selection}

Given that explanations are sampled from the LLM and we are not restricted to datasets with manually produced explanations such as e-SNLI, we select the \textbf{ANLI R3 training set}, which is one of the most difficult ones in the literature. We perform an additional filtering step by selecting only examples that were missed with a \emph{Qwen/Qwen2.5-7B-Instruct} model in an initial sweep. This is done because easy examples that are always predicted correctly produce zero-valued advantages in \cref{eq:grpo} and do not yield any update to the parameters, making them a waste of resources. Our filtered selection is available for download at HuggingFace as \hfd{pablomiralles22/ANLI-R3-one-missed}.

This filtering process could be further refined. For example, we could perform multiple sweeps with the reference model and pick those examples where the model got both a wrong and a right prediction. However, we do not focus too much on minimizing the required number of training steps to achieve peak performance. We could also perform this filtering step with the same model we are about to adapt, instead of using the same dataset from the 7B variant for all trained models. However, this would make the results less comparable.

\section{Experiments}

We thoroughly study the effect of training Large Language Models with the methodology in \cref{sec:method} for Natural Language Inference. In particular, we wish to answer the following research questions:

\begin{enumerate}[label={\bfseries RQ\arabic*)}]
    \item Is training with GRPO effective for improving performance? Are the performance variations consistent across model sizes?
    \item Do these performance gains generalize to other datasets?
    \item Does performance (both in and out of domain) improve as the number of training parameters increases? As we restrict ourselves to the use of LoRA~\cite{hu2022lora} or QLoRA~\cite{dettmers2023qlora} due to limited resources, we study the effect of increasing the rank of the update matrices.
    \item What effect does 4bit AWQ quantization~\cite{lin2024AWQActivationaware} have on model performance, both before and after GRPO training with LoRA?
    \item What does the model learn from training with this methodology?
\end{enumerate}

\paragraph{Backbone model selection}
To answer these questions, we train models in the \emph{Qwen/Qwen2.5-*B-Instruct} family with sizes 7B, 14B and 32B with LoRA or QLoRA. All models are trained with AWQ quantization due to resource constraints except the 7B model, which is trained both with and without quantization. For reference and comparison, we show in the number of parameters and model sizes for different models and LoRA ranks in \cref{tab:num-params}.

\begin{table}[!ht]
    \centering
    \caption{Number of parameters and sizes of models and adapters used in this work. Model sizes are calculated empirically by loading the models to memory.}
    \resizebox{\textwidth}{!}{\begin{tabular}{cc|cc|cc}
\toprule
Model       & LoRA rank & \# params backbone & \# params adapter & Size 16bits & Size AWQ \\
\midrule
Qwen2.5-7B  & -         & 7.615B             &                   & 14.54GB     & 5.46GB   \\
Qwen2.5-7B  & 8         & 7.615B             & 20.19M            & 14.72GB     & 5.62GB   \\
Qwen2.5-7B  & 16        & 7.615B             & 40.37M            & 14.78GB     & 5.75GB   \\
Qwen2.5-7B  & 32        & 7.615B             & 80.74M            & 15.19GB     & 6.12GB   \\
Qwen2.5-7B  & 64        & 7.615B             & 161.48M           & 15.88GB     & 6.78GB   \\
Qwen2.5-7B  & 128       & 7.615B             & 322.96M           & 17.26GB     & 7.95GB   \\
\midrule
Qwen2.5-14B & -         & 14.770B            &                   & 28.03GB     & 9.58GB   \\
Qwen2.5-14B & 64        & 14.770B            & 275.25M           & 30.54GB     & 11.72GB  \\
\midrule
Qwen2.5-32B & -         & 32.764B            &                   & 61.66GB     & 18.44GB  \\
Qwen2.5-32B & 64        & 32.764B            & 536.87M           & 66.23GB     & 22.39GB  \\
\bottomrule
\end{tabular}}
    \label{tab:num-params}
\end{table}

\paragraph{Evaluation dataset selection}
To evaluate the models, we use a suite of established NLI benchmarks that collectively test for generalization, robustness to dataset artifacts, and linguistic competence. SNLI and MultiNLI provide large-scale, diverse training and evaluation scenarios, while HANS and NLI Diagnostic target specific model heuristics and linguistic phenomena. Please note that HANS contains binary labels for entailment and non-entailment, and model predictions are transformed by mapping neutral and contradiction predictions to non-entailment ones. Finally, Counter-NLI is helpful in assessing resilience to the hypothesis-only bias. Including these datasets ensures that our evaluation captures both in-distribution and out-of-distribution performance, as well as sensitivity to known pitfalls in NLI modeling.
\section{Results and discussion}

\subsection{Model evaluation and comparison}\label{ssec:baselines}

In \cref{tab:baseline-results-1,tab:baseline-results-2} we compare the results of GRPO-trained models with other baselines.

First, we consider results from prior work. We found the best results with T5-11B~\cite{raffel2020Exploringlimits} for MultiNLI, RoBERTa-large~\cite{wang2021EntailmentFewShot} for SNLI, RoBERTa-large~\cite{wang2021infobert} for ANLI and RoBERTa-large~\cite{pmlr-v119-bras20a} for HANS and Counter NLI. We report results from \citet{kavumba2023Promptingexplanations} separately in \cref{sec:anli-sota} because our replication efforts yielded results significantly different from those originally reported.

Since all but one of the baselines from prior work are RoBERTa-large models with $355$ million parameters, we include baselines with the same backbone models as well, ensuring a fair comparison in terms of model size. 
Our internal baselines consist of: (1) a cross-encoder adaptation of the decoder model using the last token embedding for classification; (2) direct supervised fine-tuning (SFT) without chain-of-thought reasoning; (3) the base model without GRPO; and (4) the \emph{DeepSeek-R1-Distill-Qwen} series (up to 14B due to resource constraints), which are distilled for general reasoning rather than NLI specifically. Cross-encoder and SFT models are trained on the full ANLI-R3 training set, and all models are trained with a LoRA rank of 64. We discuss the results next.

\begin{table}[!ht]
    \setlength{\tabcolsep}{1pt}
    \centering
    \adjustbox{max width=\textwidth}{\begin{tabular}{rccc|ccc}
\toprule
 & \multicolumn{3}{c}{ANLI} & \multicolumn{3}{c}{HANS} \\
Training & R1 & R2 & R3 & Cons. & Lex. & Sub. \\
\midrule
\multicolumn{7}{c}{\textbf{Prior Work}} \\
\midrule
 & \databar{0.75}{0.8393333333333333}{2.6}{pastel1}{75.00\%} & \databar{0.505}{0.7513333333333333}{2.6}{pastel1}{50.50\%} & \databar{0.477}{0.7219444444444445}{2.6}{pastel1}{47.70\%} & \databar{0.385}{0.9596}{2.6}{pastel2}{38.50\%} & \databar{0.941}{0.9868333333333333}{2.6}{pastel2}{94.10\%} & \databar{0.463}{0.7832}{2.6}{pastel2}{46.30\%} \\
\midrule
\multicolumn{7}{c}{\textbf{7B}} \\
\midrule
Base & \databar{0.729}{0.8393333333333333}{2.6}{pastel1}{72.90\%} & \databar{0.629}{0.7513333333333333}{2.6}{pastel1}{62.90\%} & \databar{0.5846538782318599}{0.7219444444444445}{2.6}{pastel1}{58.47\%} & \databar{0.8794}{0.9596}{2.6}{pastel2}{87.94\%} & \databar{0.9388}{0.9868333333333333}{2.6}{pastel2}{93.88\%} & \databar{0.7245}{0.7832}{2.6}{pastel2}{72.45\%} \\
Cross-Enc. & \databar{0.7926666666666667}{0.8393333333333333}{2.6}{pastel1}{79.27\%\textsubscript{$\pm$0.31\%}} & \databar{0.6663333333333333}{0.7513333333333333}{2.6}{pastel1}{66.63\%\textsubscript{$\pm$1.12\%}} & \databar{0.6752777777777778}{0.7219444444444445}{2.6}{pastel1}{67.53\%\textsubscript{$\pm$0.64\%}} & \databar{0.7424333333333334}{0.9596}{2.6}{pastel2}{74.24\%\textsubscript{$\pm$0.36\%}} & \databar{0.9781333333333334}{0.9868333333333333}{2.6}{pastel2}{97.81\%\textsubscript{$\pm$0.65\%}} & \databar{0.6979333333333333}{0.7832}{2.6}{pastel2}{69.79\%\textsubscript{$\pm$1.51\%}} \\
SFT & \databar{0.77}{0.8393333333333333}{2.6}{pastel1}{77.00\%\textsubscript{$\pm$0.20\%}} & \databar{0.6366666666666667}{0.7513333333333333}{2.6}{pastel1}{63.67\%\textsubscript{$\pm$0.21\%}} & \databar{0.6444444444444445}{0.7219444444444445}{2.6}{pastel1}{64.44\%\textsubscript{$\pm$0.63\%}} & \databar{0.6757}{0.9596}{2.6}{pastel2}{67.57\%\textsubscript{$\pm$2.99\%}} & \databar{0.9848333333333333}{0.9868333333333333}{2.6}{pastel2}{98.48\%\textsubscript{$\pm$0.20\%}} & \databar{0.6818666666666666}{0.7832}{2.6}{pastel2}{68.19\%\textsubscript{$\pm$0.68\%}} \\
GRPO & \databar{0.7653333333333334}{0.8393333333333333}{2.6}{pastel1}{76.53\%\textsubscript{$\pm$0.76\%}} & \databar{0.6413333333333333}{0.7513333333333333}{2.6}{pastel1}{64.13\%\textsubscript{$\pm$1.25\%}} & \databar{0.6141666666666666}{0.7219444444444445}{2.6}{pastel1}{61.42\%\textsubscript{$\pm$0.58\%}} & \databar{0.8550333333333334}{0.9596}{2.6}{pastel2}{85.50\%\textsubscript{$\pm$3.31\%}} & \databar{0.9649333333333333}{0.9868333333333333}{2.6}{pastel2}{96.49\%\textsubscript{$\pm$1.11\%}} & \databar{0.7146}{0.7832}{2.6}{pastel2}{71.46\%\textsubscript{$\pm$1.43\%}} \\
R1-Distill & \databar{0.65}{0.8393333333333333}{2.6}{pastel1}{65.00\%} & \databar{0.5816649104320337}{0.7513333333333333}{2.6}{pastel1}{58.17\%} & \databar{0.4903508771929825}{0.7219444444444445}{2.6}{pastel1}{49.04\%} & \databar{0.8244}{0.9596}{2.6}{pastel2}{82.44\%} & \databar{0.8178}{0.9868333333333333}{2.6}{pastel2}{81.78\%} & \databar{0.6605}{0.7832}{2.6}{pastel2}{66.05\%} \\
\midrule
\multicolumn{7}{c}{\textbf{7B-AWQ}} \\
\midrule
Base & \databar{0.674}{0.8393333333333333}{2.6}{pastel1}{67.40\%} & \databar{0.584}{0.7513333333333333}{2.6}{pastel1}{58.40\%} & \databar{0.5358333333333334}{0.7219444444444445}{2.6}{pastel1}{53.58\%} & \databar{0.8298}{0.9596}{2.6}{pastel2}{82.98\%} & \databar{0.8878}{0.9868333333333333}{2.6}{pastel2}{88.78\%} & \databar{0.7343}{0.7832}{2.6}{pastel2}{73.43\%} \\
Cross-Enc. & \databar{0.6346666666666666}{0.8393333333333333}{2.6}{pastel1}{63.47\%\textsubscript{$\pm$0.51\%}} & \databar{0.5103333333333334}{0.7513333333333333}{2.6}{pastel1}{51.03\%\textsubscript{$\pm$0.74\%}} & \databar{0.5225}{0.7219444444444445}{2.6}{pastel1}{52.25\%\textsubscript{$\pm$1.26\%}} & \databar{0.7670333333333333}{0.9596}{2.6}{pastel2}{76.70\%\textsubscript{$\pm$3.80\%}} & \databar{0.9067333333333334}{0.9868333333333333}{2.6}{pastel2}{90.67\%\textsubscript{$\pm$5.05\%}} & \databar{0.7260666666666666}{0.7832}{2.6}{pastel2}{72.61\%\textsubscript{$\pm$3.95\%}} \\
SFT & \databar{0.7613333333333333}{0.8393333333333333}{2.6}{pastel1}{76.13\%\textsubscript{$\pm$0.12\%}} & \databar{0.631}{0.7513333333333333}{2.6}{pastel1}{63.10\%\textsubscript{$\pm$0.44\%}} & \databar{0.6483333333333333}{0.7219444444444445}{2.6}{pastel1}{64.83\%\textsubscript{$\pm$0.87\%}} & \databar{0.7364}{0.9596}{2.6}{pastel2}{73.64\%\textsubscript{$\pm$1.38\%}} & \databar{0.9848666666666667}{0.9868333333333333}{2.6}{pastel2}{98.49\%\textsubscript{$\pm$0.11\%}} & \databar{0.6913333333333332}{0.7832}{2.6}{pastel2}{69.13\%\textsubscript{$\pm$0.63\%}} \\
GRPO & \databar{0.7523333333333334}{0.8393333333333333}{2.6}{pastel1}{75.23\%\textsubscript{$\pm$1.46\%}} & \databar{0.6236666666666667}{0.7513333333333333}{2.6}{pastel1}{62.37\%\textsubscript{$\pm$1.16\%}} & \databar{0.5947222222222223}{0.7219444444444445}{2.6}{pastel1}{59.47\%\textsubscript{$\pm$0.97\%}} & \databar{0.8057333333333334}{0.9596}{2.6}{pastel2}{80.57\%\textsubscript{$\pm$3.26\%}} & \databar{0.9458333333333333}{0.9868333333333333}{2.6}{pastel2}{94.58\%\textsubscript{$\pm$2.47\%}} & \databar{0.7197333333333334}{0.7832}{2.6}{pastel2}{71.97\%\textsubscript{$\pm$1.50\%}} \\
\midrule
\multicolumn{7}{c}{\textbf{14B}} \\
\midrule
R1-Distill & \databar{0.8048780487804879}{0.8393333333333333}{2.6}{pastel1}{80.49\%} & \databar{0.7109053497942387}{0.7513333333333333}{2.6}{pastel1}{71.09\%} & \databar{0.6300085251491901}{0.7219444444444445}{2.6}{pastel1}{63.00\%} & \databar{0.9406}{0.9596}{2.6}{pastel2}{94.06\%} & \databar{0.9391}{0.9868333333333333}{2.6}{pastel2}{93.91\%} & \databar{0.709}{0.7832}{2.6}{pastel2}{70.90\%} \\
\midrule
\multicolumn{7}{c}{\textbf{14B-AWQ}} \\
\midrule
Base & \databar{0.7317317317317318}{0.8393333333333333}{2.6}{pastel1}{73.17\%} & \databar{0.6386386386386387}{0.7513333333333333}{2.6}{pastel1}{63.86\%} & \databar{0.5683333333333334}{0.7219444444444445}{2.6}{pastel1}{56.83\%} & \databar{0.93}{0.9596}{2.6}{pastel2}{93.00\%} & \databar{0.8932}{0.9868333333333333}{2.6}{pastel2}{89.32\%} & \databar{0.7832}{0.7832}{2.6}{pastel2}{\textbf{78.32\%}} \\
Cross-Enc. & \databar{0.6960000000000001}{0.8393333333333333}{2.6}{pastel1}{69.60\%\textsubscript{$\pm$0.62\%}} & \databar{0.5926666666666667}{0.7513333333333333}{2.6}{pastel1}{59.27\%\textsubscript{$\pm$1.00\%}} & \databar{0.5633333333333334}{0.7219444444444445}{2.6}{pastel1}{56.33\%\textsubscript{$\pm$1.08\%}} & \databar{0.7443333333333334}{0.9596}{2.6}{pastel2}{74.43\%\textsubscript{$\pm$4.11\%}} & \databar{0.9719333333333333}{0.9868333333333333}{2.6}{pastel2}{97.19\%\textsubscript{$\pm$1.23\%}} & \databar{0.7519}{0.7832}{2.6}{pastel2}{75.19\%\textsubscript{$\pm$2.42\%}} \\
SFT & \databar{0.8193333333333332}{0.8393333333333333}{2.6}{pastel1}{81.93\%\textsubscript{$\pm$0.71\%}} & \databar{0.6873333333333335}{0.7513333333333333}{2.6}{pastel1}{68.73\%\textsubscript{$\pm$1.27\%}} & \databar{0.6669444444444445}{0.7219444444444445}{2.6}{pastel1}{66.69\%\textsubscript{$\pm$0.79\%}} & \databar{0.9281333333333334}{0.9596}{2.6}{pastel2}{92.81\%\textsubscript{$\pm$2.65\%}} & \databar{0.9868333333333333}{0.9868333333333333}{2.6}{pastel2}{\textbf{98.68\%}\textsubscript{$\pm$0.33\%}} & \databar{0.6948666666666666}{0.7832}{2.6}{pastel2}{69.49\%\textsubscript{$\pm$0.89\%}} \\
GRPO & \databar{0.783}{0.8393333333333333}{2.6}{pastel1}{78.30\%\textsubscript{$\pm$0.53\%}} & \databar{0.6693333333333333}{0.7513333333333333}{2.6}{pastel1}{66.93\%\textsubscript{$\pm$1.47\%}} & \databar{0.6222222222222222}{0.7219444444444445}{2.6}{pastel1}{62.22\%\textsubscript{$\pm$0.77\%}} & \databar{0.9267}{0.9596}{2.6}{pastel2}{92.67\%\textsubscript{$\pm$4.31\%}} & \databar{0.9538000000000001}{0.9868333333333333}{2.6}{pastel2}{95.38\%\textsubscript{$\pm$1.25\%}} & \databar{0.7651333333333333}{0.7832}{2.6}{pastel2}{76.51\%\textsubscript{$\pm$1.99\%}} \\
\midrule
\multicolumn{7}{c}{\textbf{32B-AWQ}} \\
\midrule
Base & \databar{0.798}{0.8393333333333333}{2.6}{pastel1}{79.80\%} & \databar{0.708}{0.7513333333333333}{2.6}{pastel1}{70.80\%} & \databar{0.6608333333333334}{0.7219444444444445}{2.6}{pastel1}{66.08\%} & \databar{0.9596}{0.9596}{2.6}{pastel2}{\textbf{95.96\%}} & \databar{0.9095}{0.9868333333333333}{2.6}{pastel2}{90.95\%} & \databar{0.7597}{0.7832}{2.6}{pastel2}{75.97\%} \\
Cross-Enc. & \databar{0.7086666666666667}{0.8393333333333333}{2.6}{pastel1}{70.87\%\textsubscript{$\pm$0.38\%}} & \databar{0.6083333333333333}{0.7513333333333333}{2.6}{pastel1}{60.83\%\textsubscript{$\pm$0.90\%}} & \databar{0.5688888888888889}{0.7219444444444445}{2.6}{pastel1}{56.89\%\textsubscript{$\pm$1.13\%}} & \databar{0.6114666666666667}{0.9596}{2.6}{pastel2}{61.15\%\textsubscript{$\pm$8.85\%}} & \databar{0.8029666666666667}{0.9868333333333333}{2.6}{pastel2}{80.30\%\textsubscript{$\pm$10.96\%}} & \databar{0.5529666666666667}{0.7832}{2.6}{pastel2}{55.30\%\textsubscript{$\pm$3.84\%}} \\
SFT & \databar{0.8373333333333334}{0.8393333333333333}{2.6}{pastel1}{83.73\%\textsubscript{$\pm$0.21\%}} & \databar{0.722}{0.7513333333333333}{2.6}{pastel1}{72.20\%\textsubscript{$\pm$0.30\%}} & \databar{0.7188888888888889}{0.7219444444444445}{2.6}{pastel1}{71.89\%\textsubscript{$\pm$0.10\%}} & \databar{0.7790333333333334}{0.9596}{2.6}{pastel2}{77.90\%\textsubscript{$\pm$1.62\%}} & \databar{0.9820333333333333}{0.9868333333333333}{2.6}{pastel2}{98.20\%\textsubscript{$\pm$0.08\%}} & \databar{0.6534666666666666}{0.7832}{2.6}{pastel2}{65.35\%\textsubscript{$\pm$0.36\%}} \\
GRPO & \databar{0.8393333333333333}{0.8393333333333333}{2.6}{pastel1}{\textbf{83.93\%}\textsubscript{$\pm$0.49\%}} & \databar{0.7513333333333333}{0.7513333333333333}{2.6}{pastel1}{\textbf{75.13\%}\textsubscript{$\pm$0.21\%}} & \databar{0.7219444444444445}{0.7219444444444445}{2.6}{pastel1}{\textbf{72.19\%}\textsubscript{$\pm$0.39\%}} & \databar{0.9243}{0.9596}{2.6}{pastel2}{92.43\%\textsubscript{$\pm$4.41\%}} & \databar{0.9821333333333334}{0.9868333333333333}{2.6}{pastel2}{98.21\%\textsubscript{$\pm$0.60\%}} & \databar{0.7783333333333333}{0.7832}{2.6}{pastel2}{77.83\%\textsubscript{$\pm$1.19\%}} \\
\bottomrule
\end{tabular}}
    \caption{Comparison of different training techniques and baselines from prior work in ANLI and HANS. The backbones are from the \emph{Qwen/Qwen2.5-*B-Instruct-*} series, except models from the \emph{DeepSeek-R1-Distill-Qwen} series. All models from this work are trained with LoRA or QLora with a rank of 64. We show the best result for each dataset in bold. The mean and standard deviation across three runs is shown for trained models.}
    \label{tab:baseline-results-1}
\end{table}

\begin{table}[!ht]
    \setlength{\tabcolsep}{1pt}
    \centering
    \adjustbox{max width=\textwidth}{\begin{tabular}{rcccc|c|c|c}
\toprule
 & \multicolumn{4}{c}{Diagnostic} & SNLI & MNLI & Counter-NLI \\
Training & Know. & Logic & LS & PAS & Test & Mism. & Test \\
\midrule
\multicolumn{8}{c}{\textbf{Prior Work}} \\
\midrule
 & \databar{0.539}{0.7605633802816901}{2.6}{pastel3}{53.90\%} & \databar{0.587}{0.7857142857142857}{2.6}{pastel3}{58.70\%} & \databar{0.665}{0.8026729559748428}{2.6}{pastel3}{66.50\%} & \databar{0.699}{0.7771739130434782}{2.6}{pastel3}{69.90\%} & \databar{0.931}{0.931}{2.6}{pastel4}{\textbf{93.10\%}} & \databar{0.917}{0.917}{2.6}{pastel4}{\textbf{91.70\%}} & \databar{0.671}{0.8156249999999999}{2.6}{pastel4}{67.10\%} \\
\midrule
\multicolumn{8}{c}{\textbf{7B}} \\
\midrule
Base & \databar{0.7112676056338029}{0.7605633802816901}{2.6}{pastel3}{71.13\%} & \databar{0.7362637362637363}{0.7857142857142857}{2.6}{pastel3}{73.63\%} & \databar{0.7311320754716981}{0.8026729559748428}{2.6}{pastel3}{73.11\%} & \databar{0.75}{0.7771739130434782}{2.6}{pastel3}{75.00\%} & \databar{0.8060871335504886}{0.931}{2.6}{pastel4}{80.61\%} & \databar{0.835740439381611}{0.917}{2.6}{pastel4}{83.57\%} & \databar{0.70375}{0.8156249999999999}{2.6}{pastel4}{70.38\%} \\
Cross-Enc. & \databar{0.647887323943662}{0.7605633802816901}{2.6}{pastel3}{64.79\%\textsubscript{$\pm$2.20\%}} & \databar{0.6776556776556776}{0.7857142857142857}{2.6}{pastel3}{67.77\%\textsubscript{$\pm$0.16\%}} & \databar{0.7413522012578616}{0.8026729559748428}{2.6}{pastel3}{74.14\%\textsubscript{$\pm$0.72\%}} & \databar{0.7373188405797101}{0.7771739130434782}{2.6}{pastel3}{73.73\%\textsubscript{$\pm$0.57\%}} & \databar{0.9053338762214983}{0.931}{2.6}{pastel4}{90.53\%\textsubscript{$\pm$0.26\%}} & \databar{0.9064279902359642}{0.917}{2.6}{pastel4}{90.64\%\textsubscript{$\pm$0.20\%}} & \databar{0.7952083333333334}{0.8156249999999999}{2.6}{pastel4}{79.52\%\textsubscript{$\pm$0.19\%}} \\
SFT & \databar{0.6795774647887324}{0.7605633802816901}{2.6}{pastel3}{67.96\%\textsubscript{$\pm$0.61\%}} & \databar{0.6721611721611721}{0.7857142857142857}{2.6}{pastel3}{67.22\%\textsubscript{$\pm$0.96\%}} & \databar{0.7421383647798742}{0.8026729559748428}{2.6}{pastel3}{74.21\%\textsubscript{$\pm$0.14\%}} & \databar{0.7355072463768115}{0.7771739130434782}{2.6}{pastel3}{73.55\%\textsubscript{$\pm$0.57\%}} & \databar{0.9072000542888166}{0.931}{2.6}{pastel4}{90.72\%\textsubscript{$\pm$0.11\%}} & \databar{0.9009018171955518}{0.917}{2.6}{pastel4}{90.09\%\textsubscript{$\pm$0.24\%}} & \databar{0.7847916666666667}{0.8156249999999999}{2.6}{pastel4}{78.48\%\textsubscript{$\pm$0.13\%}} \\
GRPO & \databar{0.6995305164319249}{0.7605633802816901}{2.6}{pastel3}{69.95\%\textsubscript{$\pm$1.81\%}} & \databar{0.7197802197802198}{0.7857142857142857}{2.6}{pastel3}{71.98\%\textsubscript{$\pm$0.55\%}} & \databar{0.7209119496855346}{0.8026729559748428}{2.6}{pastel3}{72.09\%\textsubscript{$\pm$2.01\%}} & \databar{0.7653985507246377}{0.7771739130434782}{2.6}{pastel3}{76.54\%\textsubscript{$\pm$1.50\%}} & \databar{0.852426373106924}{0.931}{2.6}{pastel4}{85.24\%\textsubscript{$\pm$0.45\%}} & \databar{0.8591673447247085}{0.917}{2.6}{pastel4}{85.92\%\textsubscript{$\pm$0.64\%}} & \databar{0.7493244475713988}{0.8156249999999999}{2.6}{pastel4}{74.93\%\textsubscript{$\pm$0.87\%}} \\
R1-Distill & \databar{0.6030534351145038}{0.7605633802816901}{2.6}{pastel3}{60.31\%} & \databar{0.6646341463414634}{0.7857142857142857}{2.6}{pastel3}{66.46\%} & \databar{0.6141732283464567}{0.8026729559748428}{2.6}{pastel3}{61.42\%} & \databar{0.672514619883041}{0.7771739130434782}{2.6}{pastel3}{67.25\%} & \databar{0.594032359739441}{0.931}{2.6}{pastel4}{59.40\%} & \databar{0.6715454154111064}{0.917}{2.6}{pastel4}{67.15\%} & \databar{0.5799739921976593}{0.8156249999999999}{2.6}{pastel4}{58.00\%} \\
\midrule
\multicolumn{8}{c}{\textbf{7B-AWQ}} \\
\midrule
Base & \databar{0.6830985915492958}{0.7605633802816901}{2.6}{pastel3}{68.31\%} & \databar{0.7197802197802198}{0.7857142857142857}{2.6}{pastel3}{71.98\%} & \databar{0.7099056603773585}{0.8026729559748428}{2.6}{pastel3}{70.99\%} & \databar{0.7717391304347826}{0.7771739130434782}{2.6}{pastel3}{77.17\%} & \databar{0.7666938110749185}{0.931}{2.6}{pastel4}{76.67\%} & \databar{0.8082790886899919}{0.917}{2.6}{pastel4}{80.83\%} & \databar{0.670625}{0.8156249999999999}{2.6}{pastel4}{67.06\%} \\
Cross-Enc. & \databar{0.5903755868544601}{0.7605633802816901}{2.6}{pastel3}{59.04\%\textsubscript{$\pm$5.59\%}} & \databar{0.5659340659340659}{0.7857142857142857}{2.6}{pastel3}{56.59\%\textsubscript{$\pm$2.25\%}} & \databar{0.6367924528301887}{0.8026729559748428}{2.6}{pastel3}{63.68\%\textsubscript{$\pm$4.26\%}} & \databar{0.592391304347826}{0.7771739130434782}{2.6}{pastel3}{59.24\%\textsubscript{$\pm$5.43\%}} & \databar{0.7066707383279045}{0.931}{2.6}{pastel4}{70.67\%\textsubscript{$\pm$4.11\%}} & \databar{0.8167887171141851}{0.917}{2.6}{pastel4}{81.68\%\textsubscript{$\pm$1.37\%}} & \databar{0.629375}{0.8156249999999999}{2.6}{pastel4}{62.94\%\textsubscript{$\pm$2.76\%}} \\
SFT & \databar{0.6619718309859155}{0.7605633802816901}{2.6}{pastel3}{66.20\%\textsubscript{$\pm$0.61\%}} & \databar{0.6739926739926739}{0.7857142857142857}{2.6}{pastel3}{67.40\%\textsubscript{$\pm$0.96\%}} & \databar{0.7437106918238993}{0.8026729559748428}{2.6}{pastel3}{74.37\%\textsubscript{$\pm$0.27\%}} & \databar{0.7346014492753623}{0.7771739130434782}{2.6}{pastel3}{73.46\%\textsubscript{$\pm$0.87\%}} & \databar{0.9040105863192182}{0.931}{2.6}{pastel4}{90.40\%\textsubscript{$\pm$0.01\%}} & \databar{0.8995457011120153}{0.917}{2.6}{pastel4}{89.95\%\textsubscript{$\pm$0.20\%}} & \databar{0.7822916666666666}{0.8156249999999999}{2.6}{pastel4}{78.23\%\textsubscript{$\pm$0.32\%}} \\
GRPO & \databar{0.6678403755868545}{0.7605633802816901}{2.6}{pastel3}{66.78\%\textsubscript{$\pm$3.33\%}} & \databar{0.7124542124542125}{0.7857142857142857}{2.6}{pastel3}{71.25\%\textsubscript{$\pm$0.16\%}} & \databar{0.7248427672955975}{0.8026729559748428}{2.6}{pastel3}{72.48\%\textsubscript{$\pm$1.77\%}} & \databar{0.7309782608695651}{0.7771739130434782}{2.6}{pastel3}{73.10\%\textsubscript{$\pm$1.36\%}} & \databar{0.8634589218858929}{0.931}{2.6}{pastel4}{86.35\%\textsubscript{$\pm$0.27\%}} & \databar{0.8369948467588827}{0.917}{2.6}{pastel4}{83.70\%\textsubscript{$\pm$0.61\%}} & \databar{0.7754166666666666}{0.8156249999999999}{2.6}{pastel4}{77.54\%\textsubscript{$\pm$0.85\%}} \\
\midrule
\multicolumn{8}{c}{\textbf{14B}} \\
\midrule
R1-Distill & \databar{0.6872727272727273}{0.7605633802816901}{2.6}{pastel3}{68.73\%} & \databar{0.7784256559766763}{0.7857142857142857}{2.6}{pastel3}{77.84\%} & \databar{0.7142857142857143}{0.8026729559748428}{2.6}{pastel3}{71.43\%} & \databar{0.7554945054945055}{0.7771739130434782}{2.6}{pastel3}{75.55\%} & \databar{0.7298155737704918}{0.931}{2.6}{pastel4}{72.98\%} & \databar{0.8052055360462714}{0.917}{2.6}{pastel4}{80.52\%} & \databar{0.6950220541902962}{0.8156249999999999}{2.6}{pastel4}{69.50\%} \\
\midrule
\multicolumn{8}{c}{\textbf{14B-AWQ}} \\
\midrule
Base & \databar{0.7007042253521126}{0.7605633802816901}{2.6}{pastel3}{70.07\%} & \databar{0.7005494505494505}{0.7857142857142857}{2.6}{pastel3}{70.05\%} & \databar{0.7264150943396226}{0.8026729559748428}{2.6}{pastel3}{72.64\%} & \databar{0.7282608695652174}{0.7771739130434782}{2.6}{pastel3}{72.83\%} & \databar{0.8122644392380565}{0.931}{2.6}{pastel4}{81.23\%} & \databar{0.8243490642799024}{0.917}{2.6}{pastel4}{82.43\%} & \databar{0.7298311444652908}{0.8156249999999999}{2.6}{pastel4}{72.98\%} \\
Cross-Enc. & \databar{0.6631455399061034}{0.7605633802816901}{2.6}{pastel3}{66.31\%\textsubscript{$\pm$0.73\%}} & \databar{0.6492673992673993}{0.7857142857142857}{2.6}{pastel3}{64.93\%\textsubscript{$\pm$1.24\%}} & \databar{0.7209119496855346}{0.8026729559748428}{2.6}{pastel3}{72.09\%\textsubscript{$\pm$0.54\%}} & \databar{0.7092391304347826}{0.7771739130434782}{2.6}{pastel3}{70.92\%\textsubscript{$\pm$2.32\%}} & \databar{0.8340119435396308}{0.931}{2.6}{pastel4}{83.40\%\textsubscript{$\pm$0.70\%}} & \databar{0.8651681583943586}{0.917}{2.6}{pastel4}{86.52\%\textsubscript{$\pm$0.89\%}} & \databar{0.7287500000000001}{0.8156249999999999}{2.6}{pastel4}{72.88\%\textsubscript{$\pm$0.35\%}} \\
SFT & \databar{0.6713615023474179}{0.7605633802816901}{2.6}{pastel3}{67.14\%\textsubscript{$\pm$0.89\%}} & \databar{0.6822344322344321}{0.7857142857142857}{2.6}{pastel3}{68.22\%\textsubscript{$\pm$0.16\%}} & \databar{0.755503144654088}{0.8026729559748428}{2.6}{pastel3}{75.55\%\textsubscript{$\pm$0.68\%}} & \databar{0.7427536231884058}{0.7771739130434782}{2.6}{pastel3}{74.28\%\textsubscript{$\pm$0.57\%}} & \databar{0.9021104777415853}{0.931}{2.6}{pastel4}{90.21\%\textsubscript{$\pm$0.23\%}} & \databar{0.9082926498508272}{0.917}{2.6}{pastel4}{90.83\%\textsubscript{$\pm$0.12\%}} & \databar{0.805625}{0.8156249999999999}{2.6}{pastel4}{80.56\%\textsubscript{$\pm$0.25\%}} \\
GRPO & \databar{0.6948356807511736}{0.7605633802816901}{2.6}{pastel3}{69.48\%\textsubscript{$\pm$0.54\%}} & \databar{0.7307692307692308}{0.7857142857142857}{2.6}{pastel3}{73.08\%\textsubscript{$\pm$2.25\%}} & \databar{0.7602201257861636}{0.8026729559748428}{2.6}{pastel3}{76.02\%\textsubscript{$\pm$1.57\%}} & \databar{0.7518115942028986}{0.7771739130434782}{2.6}{pastel3}{75.18\%\textsubscript{$\pm$0.87\%}} & \databar{0.8828543213403837}{0.931}{2.6}{pastel4}{88.29\%\textsubscript{$\pm$0.68\%}} & \databar{0.8686554913928218}{0.917}{2.6}{pastel4}{86.87\%\textsubscript{$\pm$0.59\%}} & \databar{0.780625}{0.8156249999999999}{2.6}{pastel4}{78.06\%\textsubscript{$\pm$0.93\%}} \\
\midrule
\multicolumn{8}{c}{\textbf{32B-AWQ}} \\
\midrule
Base & \databar{0.7007042253521126}{0.7605633802816901}{2.6}{pastel3}{70.07\%} & \databar{0.7362637362637363}{0.7857142857142857}{2.6}{pastel3}{73.63\%} & \databar{0.7547169811320755}{0.8026729559748428}{2.6}{pastel3}{75.47\%} & \databar{0.7445652173913043}{0.7771739130434782}{2.6}{pastel3}{74.46\%} & \databar{0.8501628664495114}{0.931}{2.6}{pastel4}{85.02\%} & \databar{0.8431651749389748}{0.917}{2.6}{pastel4}{84.32\%} & \databar{0.774375}{0.8156249999999999}{2.6}{pastel4}{77.44\%} \\
Cross-Enc. & \databar{0.647887323943662}{0.7605633802816901}{2.6}{pastel3}{64.79\%\textsubscript{$\pm$1.83\%}} & \databar{0.6227106227106227}{0.7857142857142857}{2.6}{pastel3}{62.27\%\textsubscript{$\pm$2.55\%}} & \databar{0.6485849056603774}{0.8026729559748428}{2.6}{pastel3}{64.86\%\textsubscript{$\pm$0.00\%}} & \databar{0.6322463768115942}{0.7771739130434782}{2.6}{pastel3}{63.22\%\textsubscript{$\pm$1.50\%}} & \databar{0.7427728013029316}{0.931}{2.6}{pastel4}{74.28\%\textsubscript{$\pm$3.94\%}} & \databar{0.8416734472470844}{0.917}{2.6}{pastel4}{84.17\%\textsubscript{$\pm$0.81\%}} & \databar{0.6516666666666667}{0.8156249999999999}{2.6}{pastel4}{65.17\%\textsubscript{$\pm$3.34\%}} \\
SFT & \databar{0.710093896713615}{0.7605633802816901}{2.6}{pastel3}{71.01\%\textsubscript{$\pm$0.41\%}} & \databar{0.7115384615384616}{0.7857142857142857}{2.6}{pastel3}{71.15\%\textsubscript{$\pm$1.26\%}} & \databar{0.7696540880503145}{0.8026729559748428}{2.6}{pastel3}{76.97\%\textsubscript{$\pm$0.36\%}} & \databar{0.7436594202898551}{0.7771739130434782}{2.6}{pastel3}{74.37\%\textsubscript{$\pm$0.63\%}} & \databar{0.913578990228013}{0.931}{2.6}{pastel4}{91.36\%\textsubscript{$\pm$0.08\%}} & \databar{0.9168022782750205}{0.917}{2.6}{pastel4}{91.68\%\textsubscript{$\pm$0.08\%}} & \databar{0.8156249999999999}{0.8156249999999999}{2.6}{pastel4}{\textbf{81.56\%}\textsubscript{$\pm$0.33\%}} \\
GRPO & \databar{0.7605633802816901}{0.7605633802816901}{2.6}{pastel3}{\textbf{76.06\%}\textsubscript{$\pm$1.76\%}} & \databar{0.7857142857142857}{0.7857142857142857}{2.6}{pastel3}{\textbf{78.57\%}\textsubscript{$\pm$1.67\%}} & \databar{0.8026729559748428}{0.8026729559748428}{2.6}{pastel3}{\textbf{80.27\%}\textsubscript{$\pm$0.36\%}} & \databar{0.7771739130434782}{0.7771739130434782}{2.6}{pastel3}{\textbf{77.72\%}\textsubscript{$\pm$0.82\%}} & \databar{0.9023479913137894}{0.931}{2.6}{pastel4}{90.23\%\textsubscript{$\pm$0.77\%}} & \databar{0.8809330078654733}{0.917}{2.6}{pastel4}{88.09\%\textsubscript{$\pm$0.45\%}} & \databar{0.8127083333333335}{0.8156249999999999}{2.6}{pastel4}{81.27\%\textsubscript{$\pm$1.13\%}} \\
\bottomrule
\end{tabular}}
    \caption{Comparison of different training techniques and baselines from prior work in NLI-Diagnostic, SNLI, MNLI and Counter-NLI. The backbones are from the \emph{Qwen/Qwen2.5-*B-Instruct-*} series, except models from the \emph{DeepSeek-R1-Distill-Qwen} series. All models from this work are trained with LoRA or QLora with a rank of 64. We show the best result for each dataset in bold. The mean and standard deviation across three runs is shown for trained models.}
    \label{tab:baseline-results-2}
\end{table}

\paragraph{Prior work} All GRPO trained models perform much better in datasets other than SNLI and MNLI ones, generalizing much better in the difficult datasets. This is expected given the disparity in model sizes.

\paragraph{Base models} We generally observe big improvements in ANLI (in-distribution), Hans-Lex, SNLI, MNLI and Counter-NLI by training with GRPO. For the bigger 32B model, the GRPO-trained model outperforms the base model in all but one dataset, with very large gaps in many of them.

\paragraph{SFT} The comparison with models trained via SFT to predict the label without SFT is more complex. SFT at 7B performs better in ANLI (in-distribution), in the easier datasets (SNLI, MNLI and Hans-Lex) and in the harder datasets Diagnostic-LS and Counter-NLI. However, this pattern is reversed as we scale up the model, and the GRPO-trained at 32B parameters performs better in most of the adversarial datasets, with a big gap in some of them. Even at 7B, SFT has worse performance in the adversarial subsets HANS-Cons, HANS-Sub and Diagnostic-Logic.

\paragraph{Cross-encoders} For the 7B backbone we find that the Cross-encoder setting works better in ANLI (in-distribution), Hans-Lex, Diagnostic-LS, SNLI, MNLI and Counter-NLI, something similar to what we found in SFT. However, for AWQ-quantized models the results worsen a lot. This could be caused by the loss of information from compressing the model, making it harder to adapt to a completely new task setting.

\paragraph{Reasoning models} We now consider the results from the \emph{DeepSeek-R1-Distill-Qwen} series. In the 7B category, the performance is much worse in most datasets for the generic-reasoning model. In the 14B category, the results are very dataset-dependent, but the performance is better on average with our GRPO-trained models. However, we must consider that the GRPO-trained model is AWQ-quantized and much smaller. The 14B non-quantized model is even bigger than the 32B-AWQ models, with the GRPO-trained 32B-AWQ model being better in most datasets.

\paragraph{On efficiency beyond model size}
Model size is only one dimension of efficiency. Inference cost is equally important and depends on multiple factors, including the hardware used, the specific model architecture, and the average output length. The latter is particularly relevant for GRPO-trained models, which are known to generate substantially longer completions during reasoning tasks—driving up token usage and, therefore, cost. However, as discussed in \cref{ssec:training-dynamics}, our GRPO models do not exhibit this behavior. Their completion lengths remain comparable to those of the base models. Despite this, inference cost remains difficult to standardize, as it varies significantly with hardware availability and the distribution of completion lengths across tasks.

\paragraph{On data leakage}
Another major challenge in fairly comparing model performance arises from the potential for data leakage in large-scale pretraining datasets. When models are trained on massive, opaque corpora, it becomes increasingly difficult to ensure that evaluation data has not been seen during pretraining—directly or indirectly (see e.g.~\cite{zhu2024InferenceTimeDecontamination}). This undermines the validity of performance comparisons, as improvements may reflect memorization rather than genuine generalization. To mitigate this, we include comparisons with other training methods using the same backbone models, ensuring a consistent pretraining foundation and isolating the effect of the GRPO procedure itself.

\subsection{Ablation on decoding temperature}

\begin{figure}[!ht]
    \centering
    \includegraphics[width=\linewidth]{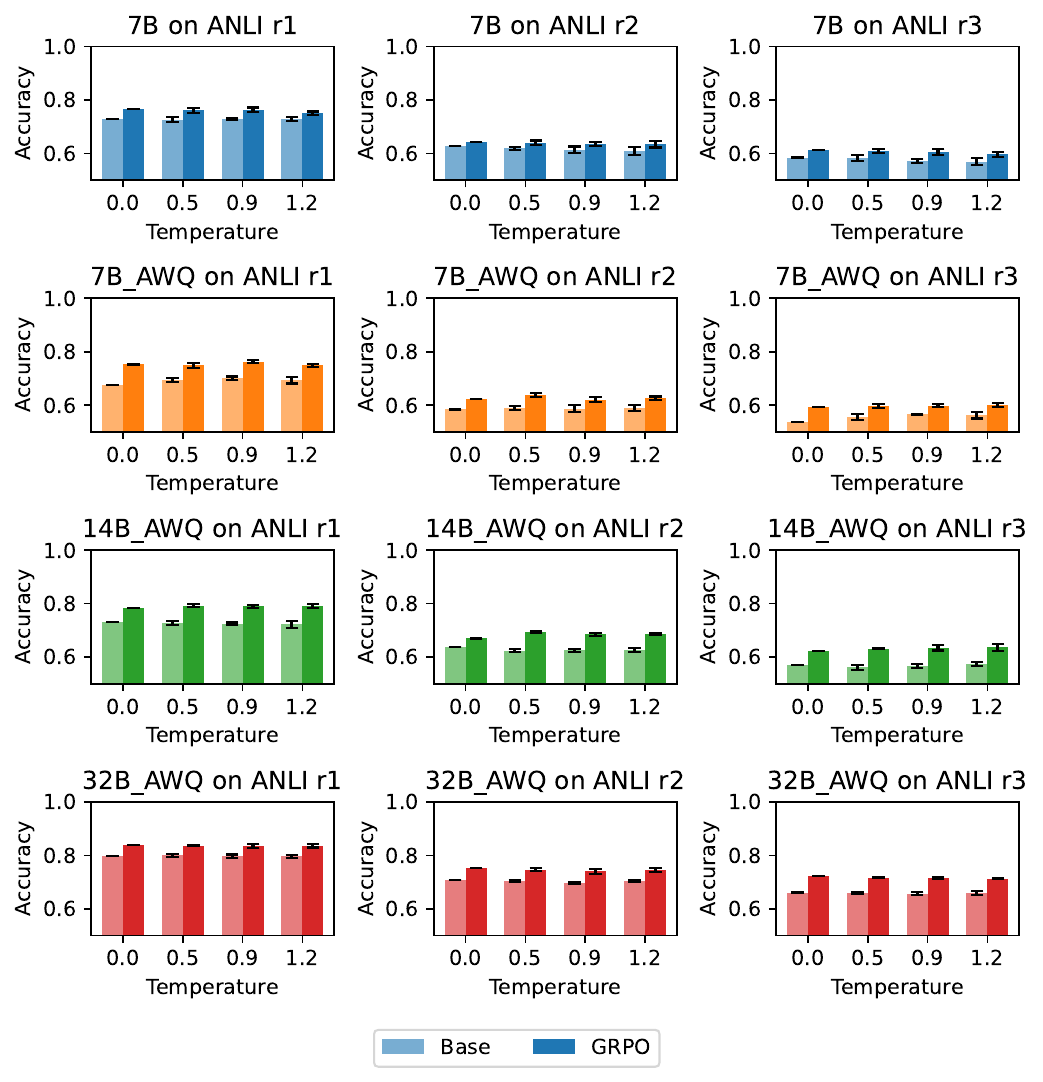}
    \caption{Effect of GRPO training on ANLI accuracy by model size, test set and decoding temperature. For non-greedy decoding, five evaluations are performed, and the standard deviation is plotted as an error line. We use models from the \lstinline{Qwen2.5-*B-Instruct} family. GRPO is performed with LoRA and a rank of $64$.}
    \label{fig:temperature}
\end{figure}

Prior to all other evaluations, we selected a decoding strategy for the rest of the experimentation. We fixed a value of $0.9$ for the \lstinline{top_p} parameter and compare different temperatures on the ANLI dataset for a selection of checkpoints. For non-greedy temperatures, we run 5 different evaluations and show the mean and standard deviation. The results are shown in \cref{fig:temperature}. We can see that it is generally equal or superior to use a greedy decoding strategy, although there is not much volatility with other temperatures. \emph{Thus, for the rest of the experimentation, including \cref{ssec:baselines}, we used greedy decoding for evaluations.} Further, we do not see any particular pattern in the performance variation across temperatures or the volatility with non-greedy temperatures between base and GRPO models.

\subsection{Performance variation across model sizes}\label{ssec:experiments-model-size}

Next, we evaluate the variation in performance across several datasets after applying our training methodology to models of different sizes. Results are shown in \cref{tab:results-by-model-size}.

As anticipated, the performance of both base and GRPO-trained models exhibits a positive correlation with model size, with larger parameter counts generally yielding improved results. The magnitude and consistency of these gains are dataset-dependent, indicating that some parts of the data distribution benefit more than others from increased model capacity.

As the model size increases, we find larger improvements in average accuracy. This is expected, as bigger models result in a larger number of trainable parameters in the LoRA update matrices. More interestingly, we find that accuracy improvements are more consistent across adversarial datasets as the size of the model increases.
The largest model present gains across all but one dataset, seemingly improving correctly and not relying on spurious correlations. On the other hand, the 7B GRPO model shows diminished results in certain adversarial datasets following fine-tuning, suggesting a possible overfitting or capacity limitation at this scale.

Finally, we highlight that the 14B GRPO model almost reaches the average performance of the base 32B model, which represents a strong efficiency gain introduced by fine-tuning with GRPO.

\begin{table}[!ht]
\centering
\caption{Increase in accuracy from GRPO training for different backbone sizes. The baseline models are from the \emph{Qwen/Qwen2.5-*B-Instruct-AWQ}, and all of them are trained with QLora and a rank of 64 and a batch size of 8.}
\label{tab:results-by-model-size}
\resizebox{\textwidth}{!}{
\begin{tabular}{cc|ccc|ccc|ccc}
\toprule
               &         & \multicolumn{3}{c}{7B (AWQ)}                        & \multicolumn{3}{c}{14B (AWQ)}                      & \multicolumn{3}{c}{32B (AWQ)}                      \\
Dataset        & Subset  & Base    & GRPO    & $\Delta$                           & Base    & GRPO    & $\Delta$                          & Base    & GRPO    & $\Delta$                          \\
\midrule
anli           & r1      & 67.40\% & 75.23\% & {\color[HTML]{02C73D} +7.83\%}  & 73.17\% & 78.30\% & {\color[HTML]{02C73D} +5.13\%} & 79.80\% & 83.93\% & {\color[HTML]{02C73D} +4.13\%} \\
anli           & r2      & 58.40\% & 62.37\% & {\color[HTML]{02C73D} +3.97\%}  & 63.86\% & 66.93\% & {\color[HTML]{02C73D} +3.07\%} & 70.80\% & 75.13\% & {\color[HTML]{02C73D} +4.33\%} \\
anli           & r3      & 53.58\% & 59.47\% & {\color[HTML]{02C73D} +5.89\%}  & 56.83\% & 62.22\% & {\color[HTML]{02C73D} +5.39\%} & 66.08\% & 72.19\% & {\color[HTML]{02C73D} +6.11\%} \\
counter-nli    &         & 67.06\% & 77.54\% & {\color[HTML]{02C73D} +10.48\%} & 72.98\% & 78.06\% & {\color[HTML]{02C73D} +5.08\%} & 77.44\% & 81.27\% & {\color[HTML]{02C73D} +3.83\%} \\
hans           & cons    & 82.98\% & 80.57\% & {\color[HTML]{FF0000} -2.41\%}  & 93.00\% & 92.67\% & {\color[HTML]{FF0000} -0.33\%} & 95.96\% & 92.43\% & {\color[HTML]{FF0000} -3.53\%} \\
hans           & lex     & 88.78\% & 94.58\% & {\color[HTML]{02C73D} +5.80\%}  & 89.32\% & 95.38\% & {\color[HTML]{02C73D} +6.06\%} & 90.95\% & 98.21\% & {\color[HTML]{02C73D} +7.26\%} \\
hans           & sub     & 73.43\% & 71.97\% & {\color[HTML]{FF0000} -1.46\%}  & 78.32\% & 76.51\% & {\color[HTML]{FF0000} -1.81\%} & 75.97\% & 77.83\% & {\color[HTML]{02C73D} +1.86\%} \\
mnli           & dev\_mm & 80.83\% & 83.70\% & {\color[HTML]{02C73D} +2.87\%}  & 82.43\% & 86.87\% & {\color[HTML]{02C73D} +4.43\%} & 84.32\% & 88.09\% & {\color[HTML]{02C73D} +3.78\%} \\
nli-diagnostic & know    & 68.31\% & 66.78\% & {\color[HTML]{FF0000} -1.53\%}  & 70.07\% & 69.48\% & {\color[HTML]{FF0000} -0.59\%} & 70.07\% & 76.06\% & {\color[HTML]{02C73D} +5.99\%} \\
nli-diagnostic & logic   & 71.98\% & 71.25\% & {\color[HTML]{FF0000} -0.73\%}  & 70.05\% & 73.08\% & {\color[HTML]{02C73D} +3.02\%} & 73.63\% & 78.57\% & {\color[HTML]{02C73D} +4.95\%} \\
nli-diagnostic & ls      & 70.99\% & 72.48\% & {\color[HTML]{02C73D} +1.49\%}  & 72.64\% & 76.02\% & {\color[HTML]{02C73D} +3.38\%} & 75.47\% & 80.27\% & {\color[HTML]{02C73D} +4.80\%} \\
nli-diagnostic & pas     & 77.17\% & 73.10\% & {\color[HTML]{FF0000} -4.08\%}  & 72.83\% & 75.18\% & {\color[HTML]{02C73D} +2.36\%} & 74.46\% & 77.72\% & {\color[HTML]{02C73D} +3.26\%} \\
snli           &         & 76.67\% & 86.35\% & {\color[HTML]{02C73D} +9.68\%}  & 81.23\% & 88.29\% & {\color[HTML]{02C73D} +7.06\%} & 85.02\% & 90.23\% & {\color[HTML]{02C73D} +5.22\%} \\
\midrule
avg.           &         & 72.12\% & 75.03\% & {\color[HTML]{02C73D} +2.91\%}  & 75.13\% & 78.38\% & {\color[HTML]{02C73D} +3.25\%} & 78.46\% & 82.46\% & {\color[HTML]{02C73D} +4.00\%} \\
\bottomrule
\end{tabular}}
\end{table}
\subsection{Effect of AWQ-quantization on performance}

Quantization greatly reduces the size of the model weights at the expense of accuracy. It is interesting to measure how much performance is lost, specially after training the LoRA update matrices. Our results are shown in \cref{tab:ablation-quantization} for 7B models and several LoRA ranks.

The base model suffers a small average drop of 2.95\% accuracy. This is not too significant considering a 3x model size reduction as seen in \cref{tab:num-params}. More interestingly, while GRPO improves average performance for both the quantized and non-quantized models, the gap between them gets smaller, specially for larger ranks. This means that we get a large improvement in computational resources without suffering in terms of performance.

\begin{table}[!ht]
    \centering
    \caption{Effect of AWQ quantization on NLI performance across several datasets and LoRA ranks for GRPO training.}
    \label{tab:ablation-quantization}
    \resizebox{\textwidth}{!}{\begin{tabular}{cc|cc|cc|cc|cc|cc|cc}
\toprule
               &         & \multicolumn{2}{c|}{Base}                 & \multicolumn{2}{c|}{Rank 8}               & \multicolumn{2}{c|}{Rank 16}              & \multicolumn{2}{c|}{Rank 32}              & \multicolumn{2}{c|}{Rank 64}              & \multicolumn{2}{c}{Rank 128}             \\
               &         & No AWQ  & $\Delta$ AWQ                      & No AWQ  & AWQ                            & No AWQ  & AWQ                            & No AWQ  & AWQ                            & No AWQ  & AWQ                            & No AWQ  & AWQ                            \\
\midrule
anli           & r1      & 72.90\% & {\color[HTML]{FF0000} -5.50\%} & 73.43\% & {\color[HTML]{FF0000} -0.87\%} & 74.87\% & {\color[HTML]{FF0000} -1.80\%} & 74.63\% & {\color[HTML]{FF0000} -0.07\%} & 76.53\% & {\color[HTML]{FF0000} -2.07\%} & 76.90\% & {\color[HTML]{FF0000} -2.20\%} \\
anli           & r2      & 62.90\% & {\color[HTML]{FF0000} -4.50\%} & 63.23\% & {\color[HTML]{FF0000} -1.47\%} & 63.40\% & {\color[HTML]{FF0000} -1.43\%} & 63.10\% & {\color[HTML]{02C73D} +0.77\%} & 64.13\% & {\color[HTML]{FF0000} -1.70\%} & 64.30\% & {\color[HTML]{FF0000} -1.53\%} \\
anli           & r3      & 58.47\% & {\color[HTML]{FF0000} -4.88\%} & 59.11\% & {\color[HTML]{FF0000} -0.58\%} & 60.25\% & {\color[HTML]{FF0000} -1.31\%} & 59.14\% & {\color[HTML]{02C73D} +0.00\%} & 61.42\% & {\color[HTML]{FF0000} -1.81\%} & 60.69\% & {\color[HTML]{FF0000} -0.97\%} \\
counter-nli    &         & 70.38\% & {\color[HTML]{FF0000} -3.31\%} & 72.27\% & {\color[HTML]{02C73D} +3.40\%} & 72.90\% & {\color[HTML]{02C73D} +3.42\%} & 73.44\% & {\color[HTML]{02C73D} +3.75\%} & 74.93\% & {\color[HTML]{02C73D} +1.73\%} & 76.23\% & {\color[HTML]{02C73D} +0.48\%} \\
hans           & cons    & 87.94\% & {\color[HTML]{FF0000} -4.96\%} & 87.79\% & {\color[HTML]{FF0000} -6.40\%} & 86.80\% & {\color[HTML]{FF0000} -6.46\%} & 87.00\% & {\color[HTML]{FF0000} -6.51\%} & 85.50\% & {\color[HTML]{02C73D} +0.33\%} & 81.98\% & {\color[HTML]{02C73D} +0.35\%} \\
hans           & lex     & 93.88\% & {\color[HTML]{FF0000} -5.10\%} & 94.33\% & {\color[HTML]{FF0000} -2.18\%} & 95.61\% & {\color[HTML]{FF0000} -2.11\%} & 95.09\% & {\color[HTML]{02C73D} +0.59\%} & 96.49\% & {\color[HTML]{FF0000} -3.64\%} & 96.34\% & {\color[HTML]{FF0000} -1.61\%} \\
hans           & sub     & 72.45\% & {\color[HTML]{02C73D} +0.98\%} & 72.67\% & {\color[HTML]{02C73D} +0.16\%} & 72.34\% & {\color[HTML]{02C73D} +0.38\%} & 71.90\% & {\color[HTML]{FF0000} -0.63\%} & 71.46\% & {\color[HTML]{02C73D} +1.44\%} & 70.99\% & {\color[HTML]{02C73D} +0.54\%} \\
mnli           & dev\_mm & 83.57\% & {\color[HTML]{FF0000} -2.75\%} & 84.29\% & {\color[HTML]{FF0000} -0.62\%} & 84.45\% & {\color[HTML]{FF0000} -1.28\%} & 84.48\% & {\color[HTML]{02C73D} +0.38\%} & 85.92\% & {\color[HTML]{FF0000} -2.60\%} & 85.81\% & {\color[HTML]{FF0000} -1.45\%} \\
nli-diagnostic & know    & 71.13\% & {\color[HTML]{FF0000} -2.82\%} & 69.48\% & {\color[HTML]{FF0000} -3.52\%} & 67.61\% & {\color[HTML]{02C73D} +0.59\%} & 70.07\% & {\color[HTML]{02C73D} +0.00\%} & 69.95\% & {\color[HTML]{FF0000} -4.11\%} & 71.13\% & {\color[HTML]{FF0000} -2.46\%} \\
nli-diagnostic & logic   & 73.63\% & {\color[HTML]{FF0000} -1.65\%} & 73.44\% & {\color[HTML]{FF0000} -4.12\%} & 73.35\% & {\color[HTML]{FF0000} -2.66\%} & 71.34\% & {\color[HTML]{02C73D} +1.56\%} & 71.98\% & {\color[HTML]{02C73D} +1.28\%} & 71.79\% & {\color[HTML]{02C73D} +1.01\%} \\
nli-diagnostic & ls      & 73.11\% & {\color[HTML]{FF0000} -2.12\%} & 72.56\% & {\color[HTML]{FF0000} -0.79\%} & 72.09\% & {\color[HTML]{FF0000} -1.52\%} & 73.11\% & {\color[HTML]{FF0000} -0.94\%} & 72.09\% & {\color[HTML]{02C73D} +0.00\%} & 73.66\% & {\color[HTML]{FF0000} -0.47\%} \\
nli-diagnostic & pas     & 75.00\% & {\color[HTML]{02C73D} +2.17\%} & 75.72\% & {\color[HTML]{FF0000} -1.81\%} & 75.18\% & {\color[HTML]{FF0000} -0.63\%} & 75.36\% & {\color[HTML]{02C73D} +0.54\%} & 76.54\% & {\color[HTML]{FF0000} -0.54\%} & 76.09\% & {\color[HTML]{FF0000} -1.72\%} \\
snli           &         & 80.61\% & {\color[HTML]{FF0000} -3.94\%} & 83.00\% & {\color[HTML]{02C73D} +1.77\%} & 84.03\% & {\color[HTML]{02C73D} +0.66\%} & 84.50\% & {\color[HTML]{02C73D} +1.70\%} & 85.24\% & {\color[HTML]{02C73D} +0.33\%} & 87.03\% & {\color[HTML]{FF0000} -1.11\%} \\
\midrule
avg.           &         & 75.07\% & {\color[HTML]{FF0000} -2.95\%} & 75.49\% & {\color[HTML]{FF0000} -1.31\%} & 75.61\% & {\color[HTML]{FF0000} -1.09\%} & 75.63\% & {\color[HTML]{02C73D} +0.09\%} & 76.32\% & {\color[HTML]{FF0000} -0.87\%} & 76.38\% & {\color[HTML]{FF0000} -0.86\%} \\
\bottomrule
\end{tabular}}
\end{table}
\subsection{Effect of LoRA rank on performance}\label{ssec:experiments-lora-rank}

We now investigate the relationship between LoRA rank and the accuracy gains obtained through GRPO training. The base model used in our experiments is \emph{Qwen/Qwen2.5-7B-Instruct} with and without quantization. Results are shown in \cref{tab:lora-rank-ablation}. We note that results from the 7B AWQ-quantized variant are different from those of \cref{tab:results-by-model-size}, as they are trained with different batch sizes.

\begin{table}[!ht]
\centering
\caption{Increase in accuracy from GRPO training by LoRA rank. The baseline model is \emph{Qwen/Qwen2.5-7B-Instruct}, both with and without quantization.}
\label{tab:lora-rank-ablation}
\resizebox{\textwidth}{!}{

\begin{tabular}{cc|c|ccccc}
\toprule
Dataset              & Subset               & No GRPO  & $\Delta$ r8                       & $\Delta$ r16                      & $\Delta$ r32                       & $\Delta$ r64                      & $\Delta$  r128                    \\
\midrule
                     &                      & \multicolumn{6}{c}{7B}                                                                                                                                                        \\
\midrule
anli                 & test\_r1             & 72.90\%  & {\color[HTML]{02C73D} +0.53\%} & {\color[HTML]{02C73D} +1.97\%} & {\color[HTML]{02C73D} +1.73\%}  & {\color[HTML]{02C73D} +3.63\%} & {\color[HTML]{02C73D} +4.00\%} \\
anli                 & test\_r2             & 62.90\%  & {\color[HTML]{02C73D} +0.33\%} & {\color[HTML]{02C73D} +0.50\%} & {\color[HTML]{02C73D} +0.20\%}  & {\color[HTML]{02C73D} +1.23\%} & {\color[HTML]{02C73D} +1.40\%} \\
anli                 & test\_r3             & 58.47\%  & {\color[HTML]{02C73D} +0.65\%} & {\color[HTML]{02C73D} +1.78\%} & {\color[HTML]{02C73D} +0.67\%}  & {\color[HTML]{02C73D} +2.95\%} & {\color[HTML]{02C73D} +2.23\%} \\
counter-nli          & test                 & 70.38\%  & {\color[HTML]{02C73D} +1.90\%} & {\color[HTML]{02C73D} +2.52\%} & {\color[HTML]{02C73D} +3.06\%}  & {\color[HTML]{02C73D} +4.56\%} & {\color[HTML]{02C73D} +5.85\%} \\
hans                 & test\_cons           & 87.94\%  & {\color[HTML]{FF0000} -0.15\%} & {\color[HTML]{FF0000} -1.14\%} & {\color[HTML]{FF0000} -0.94\%}  & {\color[HTML]{FF0000} -2.44\%} & {\color[HTML]{FF0000} -5.96\%} \\
hans                 & test\_lex            & 93.88\%  & {\color[HTML]{02C73D} +0.45\%} & {\color[HTML]{02C73D} +1.73\%} & {\color[HTML]{02C73D} +1.21\%}  & {\color[HTML]{02C73D} +2.61\%} & {\color[HTML]{02C73D} +2.46\%} \\
hans                 & test\_sub            & 72.45\%  & {\color[HTML]{02C73D} +0.22\%} & {\color[HTML]{FF0000} -0.11\%} & {\color[HTML]{FF0000} -0.55\%}  & {\color[HTML]{FF0000} -0.99\%} & {\color[HTML]{FF0000} -1.46\%} \\
mnli                 & dev\_mm              & 83.57\%  & {\color[HTML]{02C73D} +0.71\%} & {\color[HTML]{02C73D} +0.88\%} & {\color[HTML]{02C73D} +0.90\%}  & {\color[HTML]{02C73D} +2.34\%} & {\color[HTML]{02C73D} +2.24\%} \\
nli-diagnostic       & test\_know           & 71.13\%  & {\color[HTML]{FF0000} -1.64\%} & {\color[HTML]{FF0000} -3.52\%} & {\color[HTML]{FF0000} -1.06\%}  & {\color[HTML]{FF0000} -1.17\%} & {\color[HTML]{02C73D} +0.00\%} \\
nli-diagnostic       & test\_logic          & 73.63\%  & {\color[HTML]{FF0000} -0.18\%} & {\color[HTML]{FF0000} -0.27\%} & {\color[HTML]{FF0000} -2.29\%}  & {\color[HTML]{FF0000} -1.65\%} & {\color[HTML]{FF0000} -1.83\%} \\
nli-diagnostic       & test\_ls             & 73.11\%  & {\color[HTML]{FF0000} -0.55\%} & {\color[HTML]{FF0000} -1.02\%} & {\color[HTML]{02C73D} +0.00\%}  & {\color[HTML]{FF0000} -1.02\%} & {\color[HTML]{02C73D} +0.55\%} \\
nli-diagnostic       & test\_pas            & 75.00\%  & {\color[HTML]{02C73D} +0.72\%} & {\color[HTML]{02C73D} +0.18\%} & {\color[HTML]{02C73D} +0.36\%}  & {\color[HTML]{02C73D} +1.54\%} & {\color[HTML]{02C73D} +1.09\%} \\
snli                 & test                 & 80.61\%  & {\color[HTML]{02C73D} +2.40\%} & {\color[HTML]{02C73D} +3.42\%} & {\color[HTML]{02C73D} +3.89\%}  & {\color[HTML]{02C73D} +4.63\%} & {\color[HTML]{02C73D} +6.42\%} \\
\midrule
avg.                 &                      & 75.07\%  & {\color[HTML]{02C73D} +0.41\%} & {\color[HTML]{02C73D} +0.53\%} & {\color[HTML]{02C73D} +0.55\%}  & {\color[HTML]{02C73D} +1.25\%} & {\color[HTML]{02C73D} +1.31\%} \\
\midrule
                     &                      & \multicolumn{6}{c}{7B (AWQ)}                                                                                                                                                  \\
\midrule
anli                 & r1                   & 67.40\%  & {\color[HTML]{02C73D} +5.17\%} & {\color[HTML]{02C73D} +5.67\%} & {\color[HTML]{02C73D} +7.17\%}  & {\color[HTML]{02C73D} +7.07\%} & {\color[HTML]{02C73D} +7.30\%} \\
anli                 & r2                   & 58.40\%  & {\color[HTML]{02C73D} +3.37\%} & {\color[HTML]{02C73D} +3.57\%} & {\color[HTML]{02C73D} +5.47\%}  & {\color[HTML]{02C73D} +4.03\%} & {\color[HTML]{02C73D} +4.37\%} \\
anli                 & r3                   & 53.58\%  & {\color[HTML]{02C73D} +4.94\%} & {\color[HTML]{02C73D} +5.36\%} & {\color[HTML]{02C73D} +5.56\%}  & {\color[HTML]{02C73D} +6.03\%} & {\color[HTML]{02C73D} +6.14\%} \\
counter-nli          &                      & 67.06\%  & {\color[HTML]{02C73D} +8.60\%} & {\color[HTML]{02C73D} +9.25\%} & {\color[HTML]{02C73D} +10.13\%} & {\color[HTML]{02C73D} +9.60\%} & {\color[HTML]{02C73D} +9.65\%} \\
hans                 & cons                 & 82.98\%  & {\color[HTML]{FF0000} -1.59\%} & {\color[HTML]{FF0000} -2.64\%} & {\color[HTML]{FF0000} -2.50\%}  & {\color[HTML]{02C73D} +2.85\%} & {\color[HTML]{FF0000} -0.64\%} \\
hans                 & lex                  & 88.78\%  & {\color[HTML]{02C73D} +3.36\%} & {\color[HTML]{02C73D} +4.72\%} & {\color[HTML]{02C73D} +6.90\%}  & {\color[HTML]{02C73D} +4.07\%} & {\color[HTML]{02C73D} +5.95\%} \\
hans                 & sub                  & 73.43\%  & {\color[HTML]{FF0000} -0.60\%} & {\color[HTML]{FF0000} -0.71\%} & {\color[HTML]{FF0000} -2.16\%}  & {\color[HTML]{FF0000} -0.53\%} & {\color[HTML]{FF0000} -1.89\%} \\
mnli                 & dev\_mm              & 80.83\%  & {\color[HTML]{02C73D} +2.84\%} & {\color[HTML]{02C73D} +2.34\%} & {\color[HTML]{02C73D} +4.02\%}  & {\color[HTML]{02C73D} +2.49\%} & {\color[HTML]{02C73D} +3.53\%} \\
nli-diagnostic       & know                 & 68.31\%  & {\color[HTML]{FF0000} -2.35\%} & {\color[HTML]{FF0000} -0.12\%} & {\color[HTML]{02C73D} +1.76\%}  & {\color[HTML]{FF0000} -2.46\%} & {\color[HTML]{02C73D} +0.35\%} \\
nli-diagnostic       & logic                & 71.98\%  & {\color[HTML]{FF0000} -2.66\%} & {\color[HTML]{FF0000} -1.28\%} & {\color[HTML]{02C73D} +0.92\%}  & {\color[HTML]{02C73D} +1.28\%} & {\color[HTML]{02C73D} +0.82\%} \\
nli-diagnostic       & ls                   & 70.99\%  & {\color[HTML]{02C73D} +0.79\%} & {\color[HTML]{FF0000} -0.42\%} & {\color[HTML]{02C73D} +1.18\%}  & {\color[HTML]{02C73D} +1.10\%} & {\color[HTML]{02C73D} +2.20\%} \\
nli-diagnostic       & pas                  & 77.17\%  & {\color[HTML]{FF0000} -3.26\%} & {\color[HTML]{FF0000} -2.63\%} & {\color[HTML]{FF0000} -1.27\%}  & {\color[HTML]{FF0000} -1.18\%} & {\color[HTML]{FF0000} -2.81\%} \\
snli                 &                      & 76.67\%  & {\color[HTML]{02C73D} +8.11\%} & {\color[HTML]{02C73D} +8.02\%} & {\color[HTML]{02C73D} +9.53\%}  & {\color[HTML]{02C73D} +8.91\%} & {\color[HTML]{02C73D} +9.25\%} \\
\midrule
avg.                 &                      & 72.12\%  & {\color[HTML]{02C73D} +2.06\%} & {\color[HTML]{02C73D} +2.39\%} & {\color[HTML]{02C73D} +3.59\%}  & {\color[HTML]{02C73D} +3.33\%} & {\color[HTML]{02C73D} +3.40\%} \\
\bottomrule
\end{tabular}}
\end{table}

\paragraph{LoRA rank effect on non-quantized models} 
There is a clear monotonic trend in average performance as the LoRA rank increases. Among the configurations tested, LoRA rank 128 achieves the highest average accuracy across datasets. Improvement gains by increasing the number of trainable parameters is irregular, with two big jumps from the base model to a LoRA rank of 8, and from a rank of 32 to a rank of 64.

We also observe a general trend of improved performance on standard, non-adversarial benchmarks such as SNLI and MNLI as the LoRA rank increases. This indicates that higher-rank adaptations may be beneficial for non-adversarial data, as one would expect given the higher capability to fit the training data.

In contrast, the behavior on adversarial or difficult datasets is less predictable. Performance varies more significantly across ranks, with no consistent improvement pattern. We do observe that the rank with the least amount of sets with performance drops is 128, but the drop for the set targeting the constituent heuristic (i.e. assuming that a premise entails all complete subtrees in its parse tree) is significant.



\paragraph{LoRA rank effect on quantized models} For quantized models we observe an upward trend on average performance up to a rank of 32, and stability afterwards. In general, performance gains are much greater. This is not surprising, as information lost from quantization is likely easy to recover by the extra parameters. We also find that the regular performance trend for the easier datasets SNLI and MultiNLI is not present for AWQ-quantized models. 
\subsection{On training dynamics: loss, KL divergence and completion length}\label{ssec:training-dynamics}

\begin{figure}[!ht]
  \centering
  \begin{subfigure}{0.45\textwidth}
    \includegraphics[width=\linewidth]{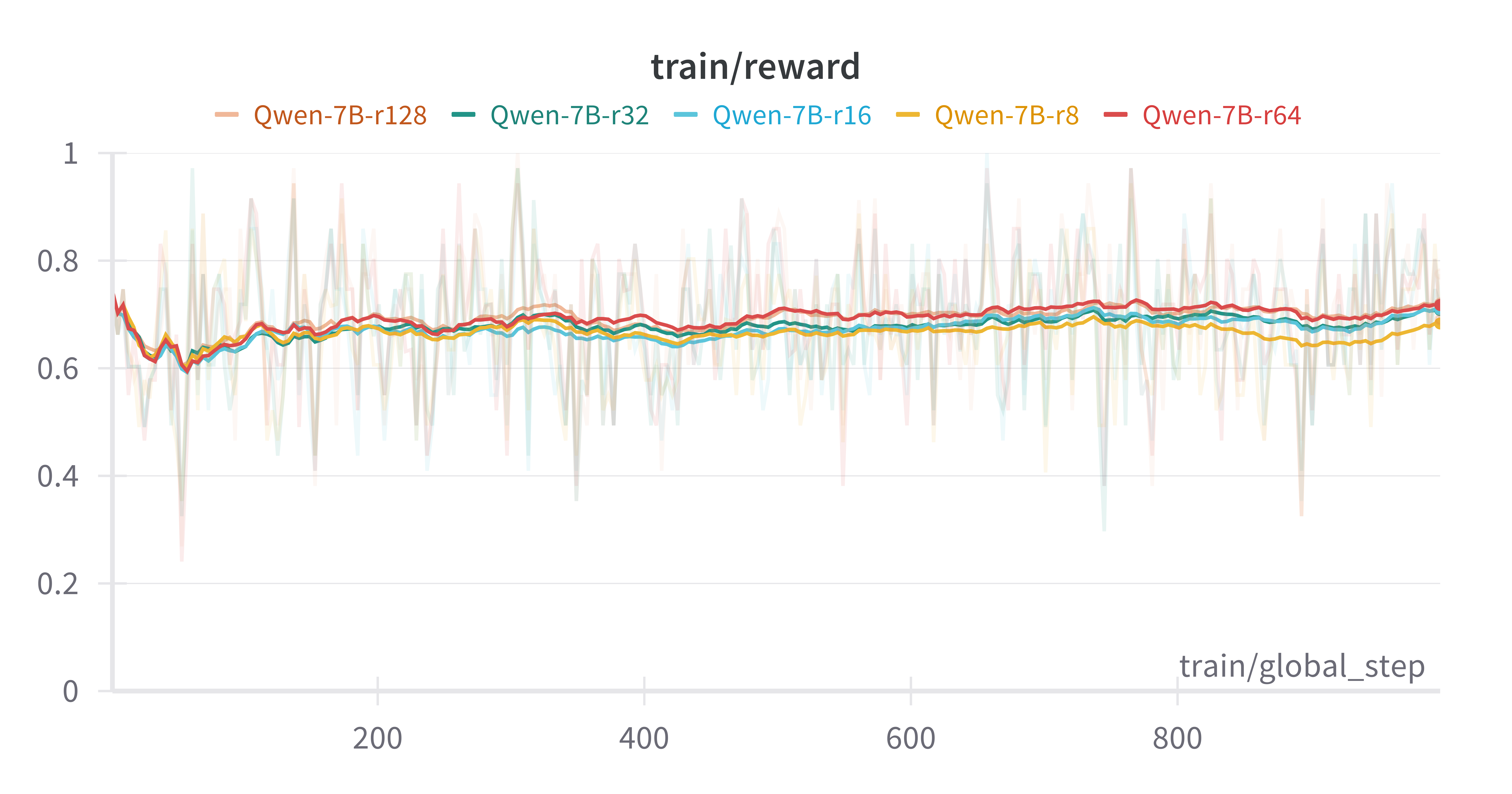}
    \caption{Reward}
    \label{subfig:training-dynamics-lora-rank-reward}
  \end{subfigure}
  \hfill
  \begin{subfigure}{0.45\textwidth}
    \includegraphics[width=\linewidth]{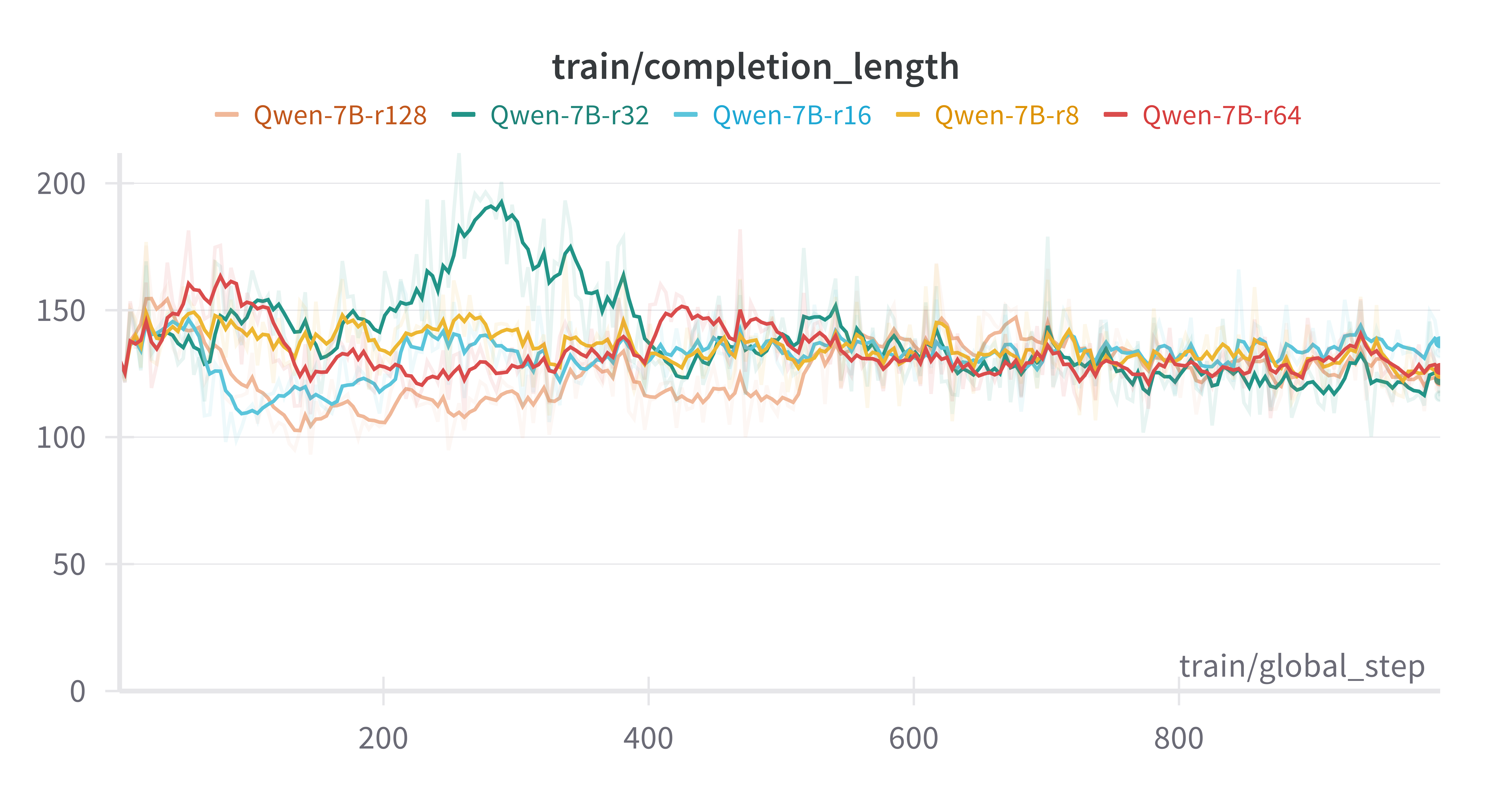}
    \caption{Completion length}
    \label{subfig:training-dynamics-lora-rank-completion-len}
  \end{subfigure}
  \hfill
  \begin{subfigure}{0.45\textwidth}
    \includegraphics[width=\linewidth]{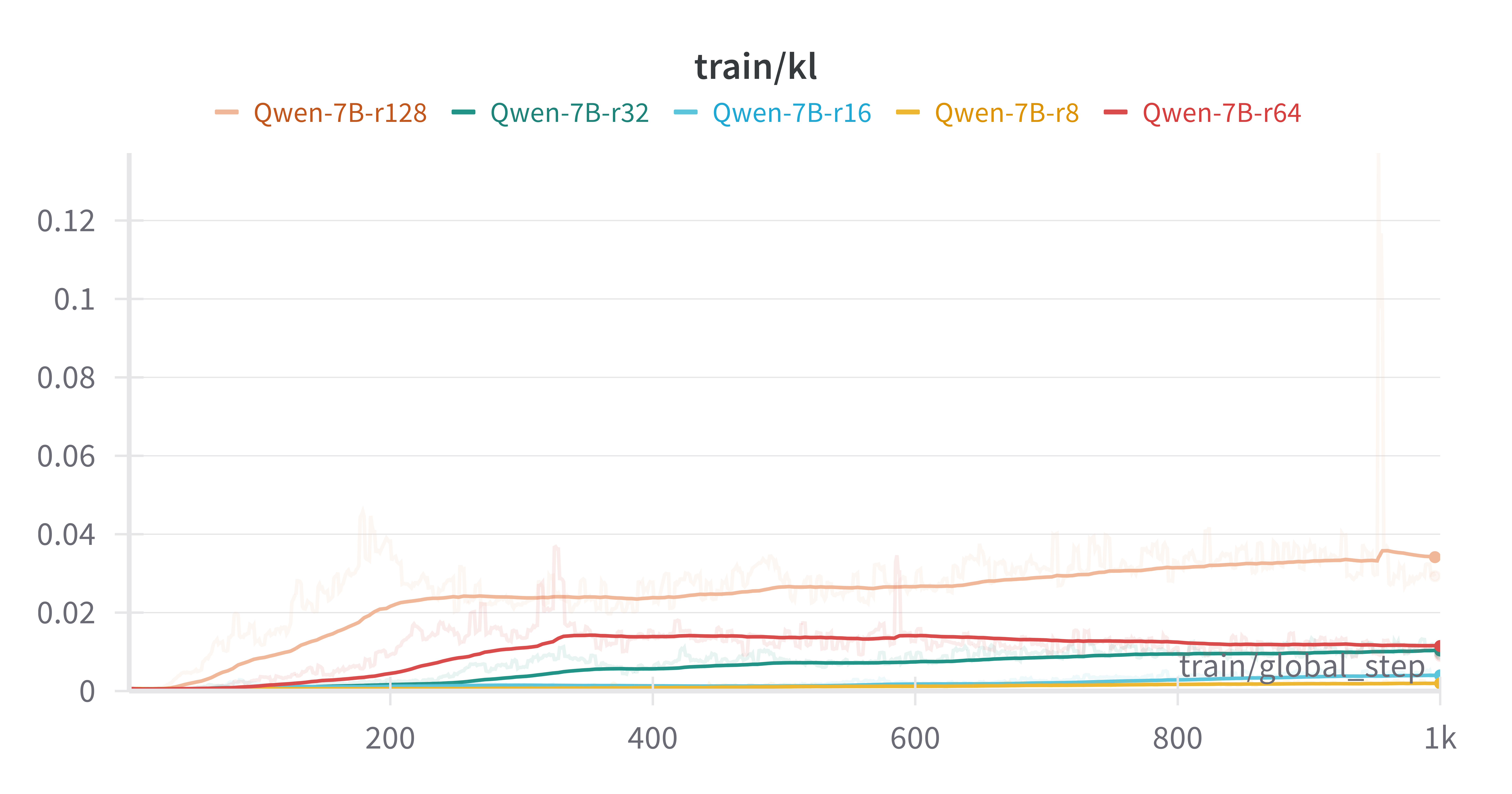}
    \caption{KL divergence}
    \label{subfig:training-dynamics-lora-rank-kl-div}
  \end{subfigure}
  \caption{Training metrics by LoRA rank for \emph{Qwen2.5-7B-Instruct}.}
  \label{fig:training-dynamics-lora-rank}
\end{figure}

\Cref{fig:training-dynamics-lora-rank} presents the training dynamics---specifically, the reward, completion length, and KL divergence---of a selection of \emph{Qwen2.5-7B-Instruct} checkpoints trained with LoRA ranks of 8, 16, 32, 64, and 128, using the GRPO algorithm. All curves are smoothed using an exponential moving average.

The final reward exhibits a slight upward trend as the LoRA rank increases, suggesting that further gains in reward may be achievable by using even higher ranks. Completion length remains largely unaffected by the training process across different ranks, showing no significant variation. This might be due to the tweaks introduce in Dr. GRPO~\cite{liu2025UnderstandingR1ZeroLike}, or a lack of sufficient parameters to introduce such a distribution shift. It is also possible that the NLI task does not benefit from longer chains of thought or retries, as evaluating an answer is the same problem as answering. KL divergence increases with LoRA rank, consistent with the expectation that a greater number of tunable parameters enables the model to diverge more from the original policy distribution.

\begin{figure}[!ht]
  \centering
  \begin{subfigure}{0.45\textwidth}
    \includegraphics[width=\linewidth]{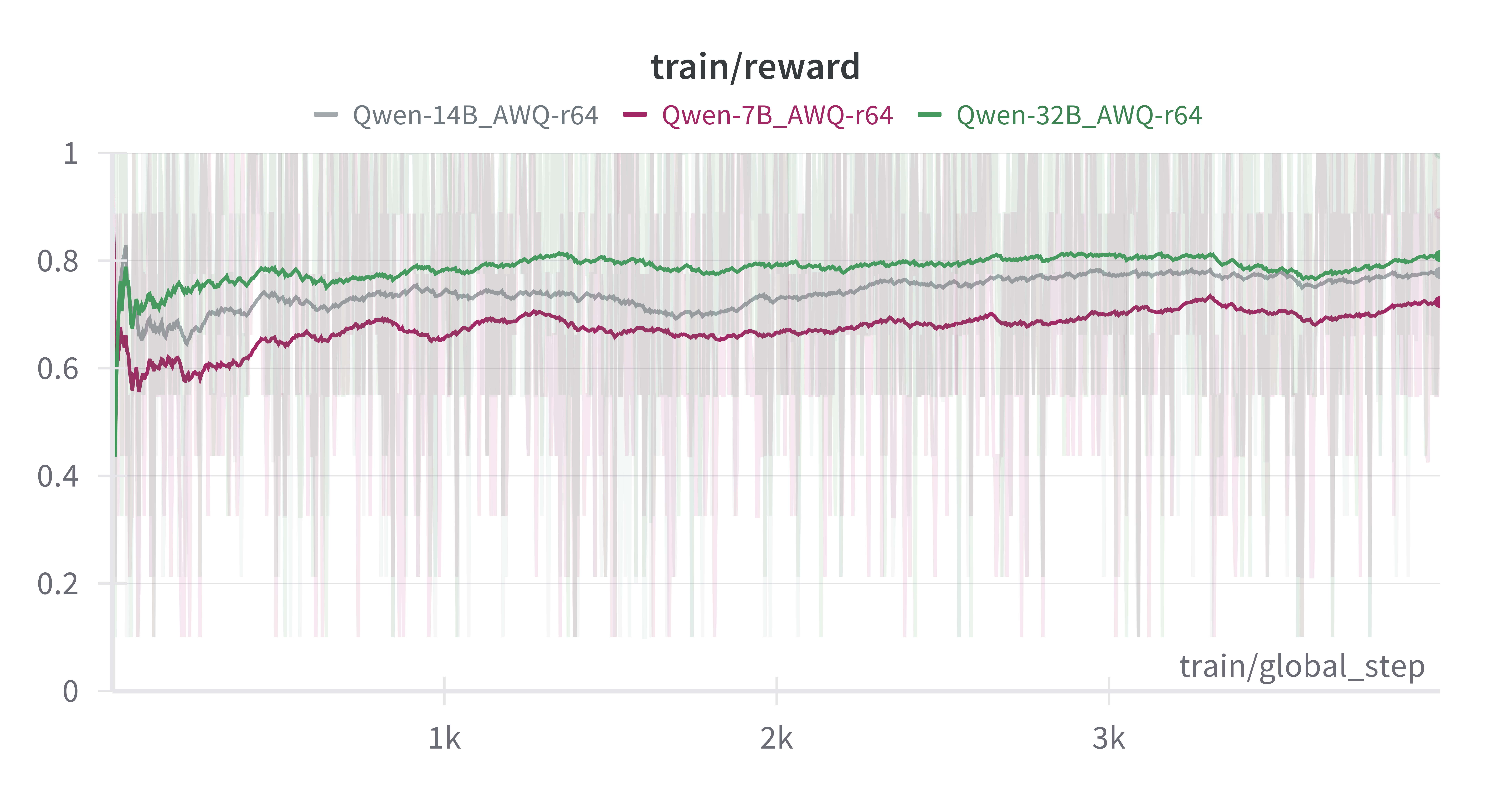}
    \caption{Reward}
    \label{subfig:training-dynamics-params-reward}
  \end{subfigure}
  \hfill
  \begin{subfigure}{0.45\textwidth}
    \includegraphics[width=\linewidth]{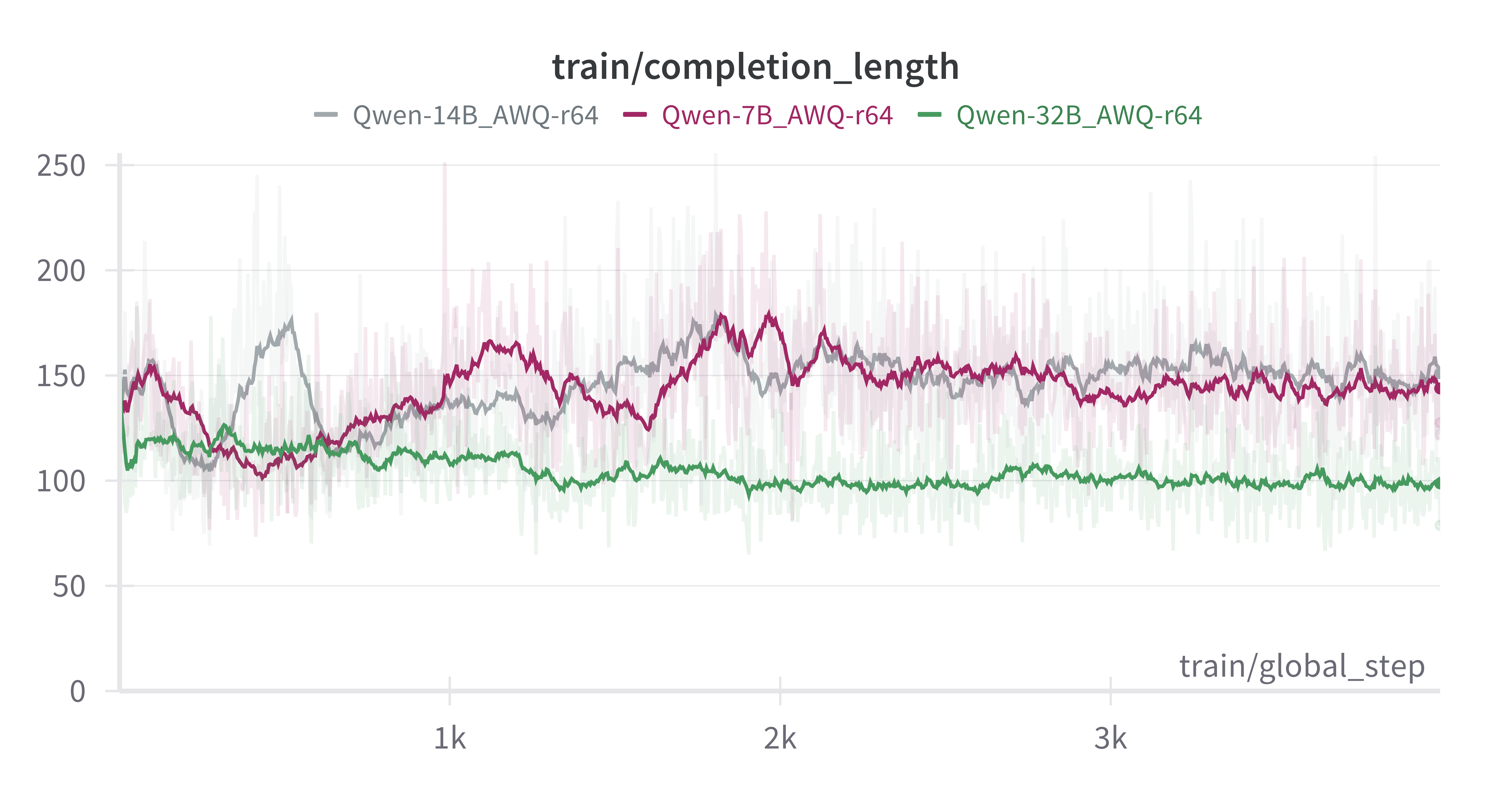}
    \caption{Completion length}
    \label{subfig:training-dynamics-params-completion-len}
  \end{subfigure}
  \hfill
  \begin{subfigure}{0.45\textwidth}
    \includegraphics[width=\linewidth]{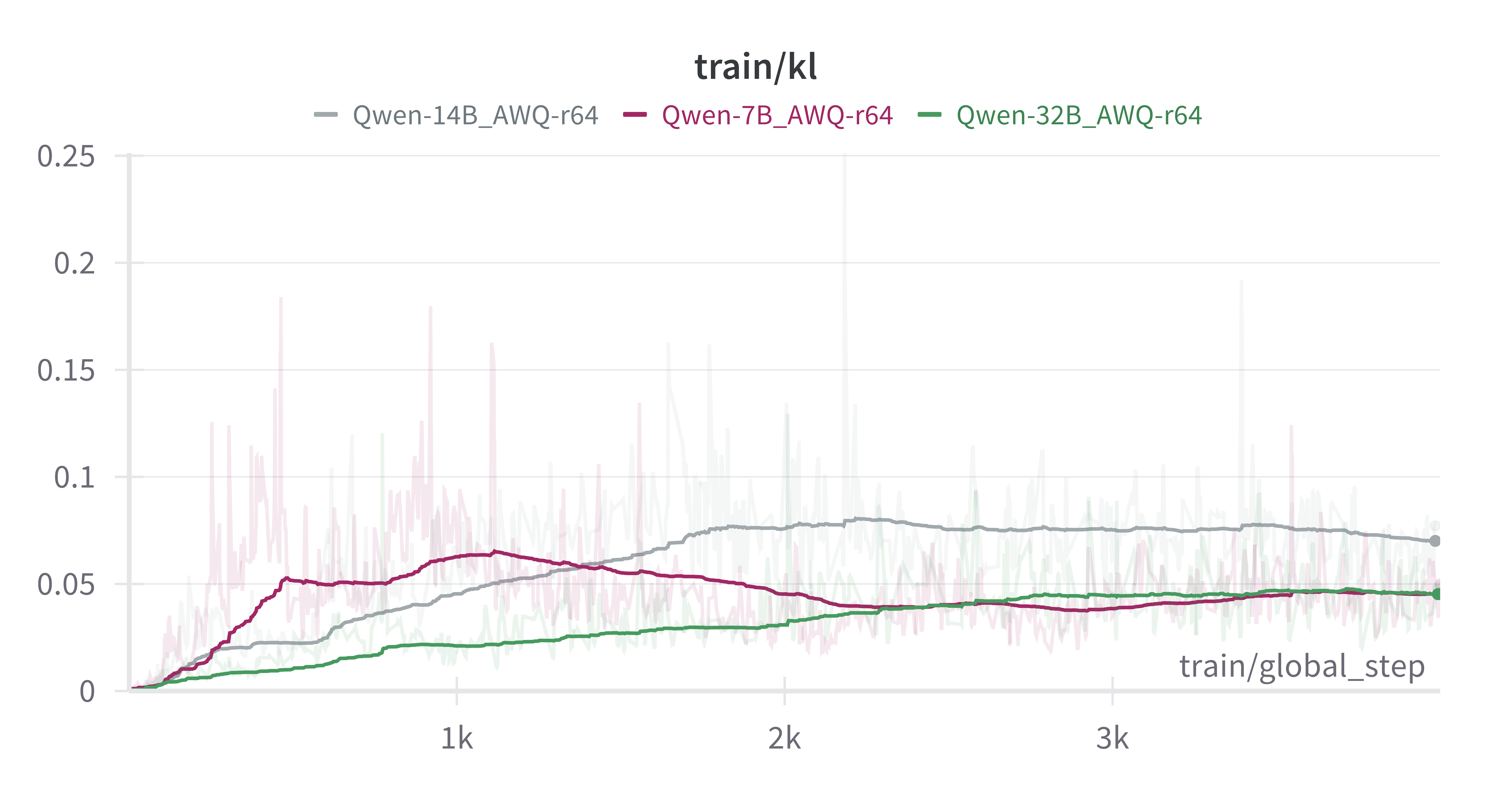}
    \caption{KL divergence}
    \label{subfig:training-dynamics-params-kl-div}
  \end{subfigure}
  \caption{Training metrics by number of parameters for \emph{Qwen2.5-*B-Instruct-AWQ} models.}
  \label{fig:training-dynamics-params}
\end{figure}

We observe similar trends for rewards in \Cref{fig:training-dynamics-params}, which displays the same training metrics across \emph{Qwen2.5-*B-Instruct-AWQ} checkpoints with varying parameter counts. Interestingly, the 7B and 14B models end with a fairly higher mean completion length than the 32B model. Of course, larger completions are only desirable if it leads to a sufficient performance gain, as it increases inference costs. Finally, we also observe a much higher final KL divergence for the 14B model.

\subsection{Analyzing differences in model outputs}\label{ssec:answer-analysis}

In this section we try to study differences in the output of models before and after training with our methodology. In particular, we consider the base model \emph{Qwen/Qwen2.5-32B-Instruct-AWQ} and one particular checkpoint of our GRPO-trained models. We restrict ourselves to examples from the ANLI R3 test set where the output of the base model and of the GRPO-trained model differ.

\begin{figure}[!ht]
  \centering
  \begin{subfigure}{0.32\textwidth}
    \includegraphics[width=\linewidth]{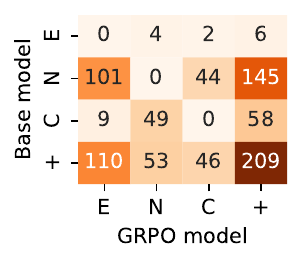}
    \caption{Base model vs. GRPO model.}
    \label{subfig:diff-model-outputs-base-grpo}
  \end{subfigure}
  \hfill
  \begin{subfigure}{0.32\textwidth}
    \includegraphics[width=\linewidth]{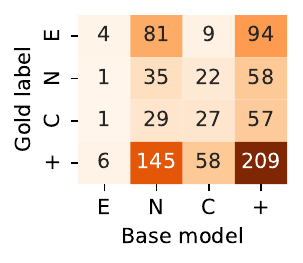}
    \caption{Gold labels vs. base model.}
    \label{subfig:diff-model-outputs-gold-base}
  \end{subfigure}
  \hfill
  \begin{subfigure}{0.32\textwidth}
    \includegraphics[width=\linewidth]{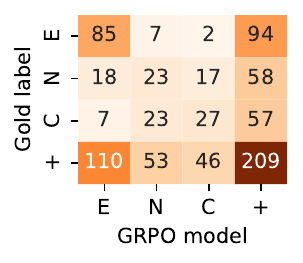}
    \caption{Gold labels vs. GRPO model.}
    \label{subfig:diff-model-outputs-gold-grpo}
  \end{subfigure}
  \caption{Confusion matrices between the base and GRPO models' predictions, and between predictions and gold labels, for ANLI R3 test examples where the base and GRPO models differ in output.}
  \label{fig:diff-model-outputs}
\end{figure}

In \cref{fig:diff-model-outputs}, we present heatmaps that display the confusion matrices comparing gold labels, base model predictions, and GRPO-trained model predictions. The most frequent pattern is that the base model predicts neutral, while the GRPO-trained model predicts entailment. We also observe several instances of neutral-contradiction and contradiction-neutral prediction pairs, which can be attributed to the subtle differences between these categories. Comparing with the gold labels, it is clear that the base model often misclassifies entailment as neutral, a mistake that the GRPO training appears to address.

In \ref{sec:output-examples} we include seven random samples from the ANLI R3 test set where \emph{Qwen/Qwen2.5-32B-Instruct-AWQ} and its GRPO-trained counterpart differ in opinion. We show the example and output of both models, as well as our analysis of the mistakes of each model.

We observe no significant difference in response patterns between the base model and the GRPO-trained version. However, we note discrepancies in the models' interpretations of word meanings, such as the distinction between ``optional'' in the first example and the nuances of ``manslaughter'' versus ``murder'' in the sixth example. Additionally, there are differences in the models' understanding of contradictions and neutrality, as seen in the second example. Furthermore, some of the gold labels appear to be ambiguous or potentially incorrect, which could contribute to the observed performance drop between validation and test datasets.
\subsection{On reward hacking}\label{ssec:reward-hacking}

Reward hacking is a phenomenon in Reinforcement Learning (RL) where an agent finds a clever, unintended way to maximize its cumulative reward without actually performing the task the designer intended.

In reasoning problems, one concern is that the model can learn to ignore the rationale provided before the answer, and predict directly from the prompt. However, we do not suspect that his behavior is occurring in our system. First, we do not see very big and sudden jumps in reward or KL-divergence in our training dynamics (\cref{ssec:training-dynamics}). Second, if this was the case, we would expect GRPO to have similar performance to training methods without CoT (like SFT in \cref{ssec:baselines}), but this is not the case. Third, in our manual analysis of answers (\cref{ssec:answer-analysis}) we do not find any outstanding disparity between the provided rationale in the final answer.

\section{Conclusions}\label{sec:conclusions}

In this work, we explored the use of reinforcement learning---specifically the Group Relative Policy Optimization (GRPO) algorithm---in conjunction with Low-Rank Adaptation (LoRA) for fine-tuning large language models (LLMs) on natural language inference (NLI) tasks. Our approach enables the training of Chain-of-Thought (CoT) reasoning without relying on human-labeled explanations, addressing a limitation in the scalability of supervised methods for NLI.

We show that GRPO training with LoRA not only improves performance on standard benchmarks such as SNLI and MultiNLI, but also on adversarial datasets where performance tends to drop in prior work. This effect is especially notable in quantized models, where LoRA serves as an effective mechanism to recover performance lost due to aggressive parameter reduction. This results in models that are both computationally efficient and capable of high-quality inference, making them practical for deployment in memory-constrained environments.

Base models without training already offer strong performance on adversarial datasets, showing stronger generalization than prior work with trained but smaller models.
This highlights the importance of starting from a sufficiently capable base model.  Indeed, if the model has not acquired the proper domain knowledge and logic capabilities from pre-training, it is very unlikely that it will be able to do so from supervised training on scarce human-generated data, which often contains artificial correlations. This is specially the case if we want our model to generalize to real-world scenarios. While the use of large-scale synthetic datasets is a potential alternative, it remains unclear whether such data can consistently provide the diversity and quality of reasoning signals needed to train smaller models effectively without introducing similar biases.

We also compare GRPO to other training methods with the same base models.
Our experiments demonstrate that our GRPO-trained models offer advantages over standard baselines, particularly as model size increases and tasks become more complex. While supervised fine-tuning (SFT) without CoT and cross-encoders perform well on standard, in-distribution datasets at smaller scales (7B), our GRPO approach scales better, outperforming SFT on most adversarial datasets at the 32B scale, with large gaps in some of them. Finally, our task-specific GRPO models consistently outperform general-purpose reasoning models at the 7B-32B scale.


\subsubsection*{Acknowledgments}
This work has been supported by the following projects: H2020 TMA-MSCA-DN TUAI project ``Towards an Understanding of Artificial Intelligence via a transparent, open and explainable perspective'' (HORIZON-MSCA-2023-DN-01-01, Grant agreement nº: 101168344); by Strategic Networking and Development Program funded by the Ministry of Science and ICT through the National Research Foundation of Korea (RS-2023-00267476); by project PCI2022-134990-2 (MARTINI) of the CHISTERA IV Cofund 2021 program; by European Comission under IBERIFIER Plus - Iberian Digital Media Observatory (DIGITAL-2023-DEPLOY- 04-EDMO-HUBS 101158511); by EMIF managed by the Calouste Gulbenkian Foundation, in the project MuseAI; and by Comunidad Autonoma de Madrid, CIRMA-CAM Project (TEC-2024/COM-404).

\printbibliography

\clearpage
\appendix
\section{Hardware and software}\label{sec:hardware}

All experiments were run on 2-3 NVIDIA GeForce RTX 3090 GPU with 24GB of memory or an NVIDIA A40 with 48GB of memory. We used PyTorch 2.6.0 with CUDA 12.4. The code for the experiments is available at \url{\repourl}. A Conda environment file can be found in the repository to reproduce the Python environment.

\section{Other training hyperparameters}\label{sec:other-hyperparams}

In \cref{tab:other-hyperparams} we show the training hyperparameters used for the SFT (asking the model to only output the label prediction) and the Cross-Encoder (using the model as a classifier via the final embedding) baselines. All models are trained on ANLI-R3.

\begin{table}[!ht]
    \centering
    \begin{tabular}{r|cc}
\toprule
              & SFT              & Cross-encoder    \\
\midrule
Learning rate           & $5\cdot 10^{-5}$ & $5\cdot 10^{-5}$ \\
Weight decay            & 0.01             & 0.01             \\
Max grad norm & 1                & 1                \\
Batch size            & 32               & 32               \\
LoRA Rank     & 64               & 64               \\
Precision     & 16b              & 16b              \\
Epochs        & 2                & 1                \\
WarmUp Steps  & 50               & 50               \\
\bottomrule
\end{tabular}

    \caption{Training hyperparameters for the SFT and Cross-Encoder baselines.}
    \label{tab:other-hyperparams}
\end{table}
\section{On the current state-of-the-art model}\label{sec:anli-sota}

As the results from~\cite{kavumba2023Promptingexplanations} seem strong for a 3B model, and no code has been made publicly available yet, we decided to replicate their experiments ourselves. The code for this replication is found in our repository, in the script \lstinline{scripts/encoder_decoder/train.py}. Examples of launching training and testing are found in the \lstinline{README.md}. Results are shown in \cref{tab:sota-replicated}, and as we see, we get significantly less performance.

\begin{table}[!ht]
\centering
\caption{Comparison between results reported in~\cite{kavumba2023Promptingexplanations} and our replication of their experiments. We also report the results of our best model.}
\label{tab:sota-replicated}
\begin{tabular}{cc|cc|c}
\toprule
Dataset        & Subset & Reported         & Replicated    & 32B (AWQ, r64)   \\
\midrule
anli           & r1     & {\ul 81.80\%}    & 52.10\%       & \textbf{83.93\%} \\
anli           & r2     & {\ul 72.50\%}    & 39.70\%       & \textbf{75.13\%} \\
anli           & r3     & \textbf{74.80\%} & 35.42\%       & {\ul 72.19\%}    \\
counter-nli    &        & \textbf{88.30\%} & 79.00\%       & {\ul 81.27\%}    \\
hans           & cons   & 17.10\%          & {\ul 54.21\%} & \textbf{92.43\%} \\
hans           & lex    & {\ul 94.20\%}    & 91.31\%       & \textbf{98.21\%} \\
hans           & sub    & 30.30\%          & {\ul 55.77\%} & \textbf{77.83\%} \\
nli-diagnostic & know   & \textbf{76.40\%} & 58.45\%       & {\ul 76.06\%}    \\
nli-diagnostic & logic  & {\ul 73.90\%}    & 57.97\%       & \textbf{78.57\%} \\
nli-diagnostic & ls     & {\ul 79.30\%}    & 69.58\%       & \textbf{80.27\%} \\
nli-diagnostic & pas    & \textbf{80.90\%} & 67.94\%       & {\ul 77.72\%}    \\
snli           &        & \textbf{95.10\%} & {\ul 92.81\%} & 90.23\%          \\
\bottomrule
\end{tabular}
\end{table}
\section{Artifacts}

\subsection{Models}

We upload one of our LoRA adapters for each base model to HuggingFace. From the non-quantized 7B model:
\begin{itemize}
    \item \hf{pablomiralles22/Qwen2.5-7B-GRPO\_r8-NLI}
    \item \hf{pablomiralles22/Qwen2.5-7B-GRPO\_r16-NLI}
    \item \hf{pablomiralles22/Qwen2.5-7B-GRPO\_r32-NLI}
    \item \hf{pablomiralles22/Qwen2.5-7B-GRPO\_r64-NLI}
    \item \hf{pablomiralles22/Qwen2.5-7B-GRPO\_r128-NLI}
    
    \item \hf{pablomiralles22/Qwen2.5-7B\_AWQ-GRPO\_r8-NLI}
    \item \hf{pablomiralles22/Qwen2.5-7B\_AWQ-GRPO\_r16-NLI}
    \item \hf{pablomiralles22/Qwen2.5-7B\_AWQ-GRPO\_r32-NLI}
    \item \hf{pablomiralles22/Qwen2.5-7B\_AWQ-GRPO\_r64\_bs32-NLI} (trained with a batch size of 32)
    \item \hf{pablomiralles22/Qwen2.5-7B\_AWQ-GRPO\_r128-NLI}
\end{itemize}

From the quantized 7B, 14B and 32B models:
\begin{itemize}
    \item \hf{pablomiralles22/Qwen2.5-7B-AWQ-GRPO\_r64-NLI} (trained with a batch size of 8)
    \item \hf{pablomiralles22/Qwen2.5-14B-AWQ-GRPO\_r64-NLI}
    \item \hf{pablomiralles22/Qwen2.5-32B-AWQ-GRPO\_r64-NLI}
\end{itemize}

\subsection{Datasets}

We create a standard format for all used datasets. In particular, we parse and upload to HuggingFace the following ones:
\hfd{pablomiralles22/esnli}, \hfd{pablomiralles22/hans}, \hfd{pablomiralles22/counter-nli}, and \hfd{pablomiralles22/nli-diagnostic}. The code used to generate them is found in the source repository, in \lstinline{notebooks/datasets.ipynb}. We also upload the GRPO training dataset at \hfd{pablomiralles22/ANLI-R3-one-missed}.
\section{Prompt template}\label{sec:prompts}
\begin{lstlisting}
|{\color{Cyan}[SYSTEM]}| Respond in the following format:
<think>...</think>
<answer>contradiction/entailment/neutral</answer>
|{\color{Cyan}[USER]}| You are given a Natural Language Inference (NLI) task. You are given a premise and a hypothesis, and you need to determine whether the hypothesis is a contradiction, entailment, or neutral given the premise.
Premise: |{\color{Lavender}\{premise\}}|
Hypothesis: |{\color{Lavender}\{hypothesis\}}|
Given the premise, is the hypothesis a contradiction, entailment, or neutral? Think step-by-step.
Use the following format for your response:
<think>...</think>
<answer>contradiction/entailment/neutral</answer>
\end{lstlisting}
\section{Model output examples}\label{sec:output-examples}

The following are seven examples from the Round 3 test set in ANLI, with predictions from \emph{Qwen/Qwen2.5-32B-Instruct-AWQ} and its GRPO-trained counterpart. We have exclusively selected samples where the base and fine-tuned models predict differently.

\noindent\makebox[\linewidth]{\rule{\linewidth}{0.4pt}} 
\begin{center} Example 1 \end{center}

\begin{lstlisting}
|{\color{Cyan}Premise}|: How to create blue roses<br>Gather the necessary materials. To dye the petals through absorption, you will need powdered blue florist's dye, water, freshly cut white, scissors, dethorning tool (optional), a glass vase, and flower preservative (optional). Ideally, the roses will not have been soaked in or treated with water. 
|{\color{Cyan}Hypothesis}|: the dethorning tool isn't needed for blue roses. 
|{\color{Cyan}True label}|: entailment 

|{\color{Cyan}Predicted answer (Base)}|: neutral 
|{\color{Cyan}Output (Base)}|: <think>The premise lists the dethorning tool as optional among the necessary materials for creating blue roses. The hypothesis states that the dethorning tool isn't needed for blue roses. Since the dethorning tool is listed as optional, it implies that it may or may not be needed, which aligns with the hypothesis. Therefore, the hypothesis is not in contradiction with the premise, nor does it follow as an entailment since the tool is optional and not definitively needed or not needed.</think>
<answer>neutral</answer> 

|{\color{Cyan}Predicted answer (GRPO)}|: entailment 
|{\color{Cyan}Output (GRPO)}|: <think>The premise states that a dethorning tool is optional when creating blue roses. This implies that the dethorning tool is not strictly necessary, which aligns with the hypothesis that the dethorning tool isn't needed for blue roses. Therefore, the hypothesis is supported by the information given in the premise.</think>
<answer>entailment</answer> 

|{\color{Cyan}Analysis}|: The base model correctly understands that ``optional'' does not imply that the action is ``not needed''; it only implies it is not mandatory. The GRPO model incorrectly assumes that ``optional'' entails ``not needed'', which is logically invalid.
\end{lstlisting}

\noindent\makebox[\linewidth]{\rule{\linewidth}{0.4pt}} 
\begin{center} Example 2 \end{center}

\begin{lstlisting}
|{\color{Cyan}Premise}|: Gregory Amenoff (born 1948) is an American painter. He is located in the tradition of the early American Modernist painters Georgia O'Keeffe, Charles Burchfield, Milton Avery, Arthur Dove and Marsden Hartley. In the early 80s his work was often associated with a style of painting called organic abstraction and exhibited alongside artists Bill Jensen, Katherine Porter and Terry Winters. 
|{\color{Cyan}Hypothesis}|: Gregory Amenoff played baseball. 
|{\color{Cyan}True label}|: contradiction 

|{\color{Cyan}Predicted answer (Base)}|: contradiction 
|{\color{Cyan}Output (Base)}|: <think>The premise provides information about Gregory Amenoff being an American painter and his association with certain artistic styles and other artists. There is no mention of him playing baseball or any sports. The hypothesis states that Gregory Amenoff played baseball, which is not supported by the information given in the premise. Since the premise is about Amenoff's career as a painter and does not mention anything about him playing baseball, the hypothesis is not entailed by the premise. Moreover, the hypothesis contradicts the focus of the premise, which is on Amenoff's career as a painter. Therefore, the hypothesis is a contradiction given the premise.</think>
<answer>contradiction</answer> 

|{\color{Cyan}Predicted answer (GRPO)}|: neutral 
|{\color{Cyan}Output (GRPO)}|: <think>The premise provides information about Gregory Amenoff being an American painter and his association with certain art styles and artists. There is no mention of him playing baseball. The hypothesis introduces new information not covered in the premise, and there is no logical connection between being a painter and playing baseball. Therefore, the hypothesis is not supported by the premise, and we cannot conclude that it is true based on the given information.</think>
<answer>neutral</answer> 

|{\color{Cyan}Analysis}|: The premise describes an American painter and his work, while the hypothesis claims the man ``played basketball.'' The gold label is contradiction, but the GRPO model classifies it as neutral. Given that the premise does not mention any specific actions, it does not explicitly exclude the possibility that the man also played basketball. Thus, the hypothesis is not directly contradicted. We agree with the GRPO model's neutral classification in this case.
\end{lstlisting}

\noindent\makebox[\linewidth]{\rule{\linewidth}{0.4pt}} 
\begin{center} Example 3 \end{center}

\begin{lstlisting}
|{\color{Cyan}Premise}|: ZURICH/BERLIN, Jan 29 (Reuters) - Here are some of the main factors that may affect Swiss stocks on Tuesday: COMPANY STATEMENTS * Gurit Holding said it will enlarge its group executive committee to eight members, from six at present, effective Feb. 1. * LafargeHolcim said it was exploring options for its business in the Philippines, including a potential sale, as the cement giant seeks to further reduce debt by selling non-core assets. ECONOMY Swiss December trade data due at 0700 GMT. (Reporting by Zurich newsroom and Berlin Speed Desk) 
|{\color{Cyan}Hypothesis}|: Gurit Holding was born Jan 29 
|{\color{Cyan}True label}|: contradiction 

|{\color{Cyan}Predicted answer (Base)}|: neutral 
|{\color{Cyan}Output (Base)}|: <think>The premise provides information about Swiss stocks and company statements, including one about Gurit Holding. The hypothesis states that Gurit Holding was born on Jan 29. The premise does not provide any information about the founding date of Gurit Holding. It only mentions that Gurit Holding will enlarge its group executive committee. Therefore, the hypothesis is not supported by the premise, nor does it contradict the premise. There is not enough information to determine if the hypothesis is true or false based on the premise.</think>
<answer>neutral</answer> 

|{\color{Cyan}Predicted answer (GRPO)}|: contradiction 
|{\color{Cyan}Output (GRPO)}|: <think>The premise states that the information is from ZURICH/BERLIN, Jan 29 (Reuters), and it mentions that Gurit Holding said it will enlarge its group executive committee. However, there is no information about when Gurit Holding was born or founded. The date Jan 29 in the premise refers to the date of the report, not the birth or founding date of Gurit Holding.</think>
<answer>contradiction</answer> 

|{\color{Cyan}Analysis}|: The premise is a news article dated January 29 that discusses Swiss stocks and mentions a specific company. The hypothesis claims that the company was founded on January 29. The gold label is contradiction, but this labeling seems incorrect. The premise does not provide any information about the company's founding date, so there is no direct contradiction. The founding date could coincide with the date of the article without conflicting with the provided information.
Therefore, the hypothesis should be labeled neutral. The base model is correct in this instance, while the GRPO model is incorrect in assuming that the overlap in dates implies contradiction.
\end{lstlisting}

\noindent\makebox[\linewidth]{\rule{\linewidth}{0.4pt}} 
\begin{center} Example 4 \end{center}

\begin{lstlisting}
|{\color{Cyan}Premise}|: How to drink whiskey<br>Pour " two fingers " of whiskey into a rocks or tulip glass. Rocks glasses, the classic whiskey cup, are short, round glasses made for 12-14 oz or liquid. Tulip glasses are curved wide at the bottom and thinner up top, concentrating the smells near your nose, and are used for more high-class whiskey tastings. 
|{\color{Cyan}Hypothesis}|: Rock glasses are a set size.  
|{\color{Cyan}True label}|: entailment 

|{\color{Cyan}Predicted answer (Base)}|: neutral 
|{\color{Cyan}Output (Base)}|: <think>The premise states that rocks glasses are "short, round glasses made for 12-14 oz of liquid." This implies that rocks glasses are typically within a certain size range but does not explicitly state that they are a set size. The hypothesis claims that "rock glasses are a set size," which is not directly supported or contradicted by the premise. The premise provides a range rather than a single set size, so the hypothesis cannot be confirmed as true based on the information given. Therefore, the relationship between the premise and the hypothesis is neutral.</think>
<answer>neutral</answer> 

|{\color{Cyan}Predicted answer (GRPO)}|: entailment 
|{\color{Cyan}Output (GRPO)}|: <think>The premise states that rocks glasses are "short, round glasses made for 12-14 oz of liquid." This implies that rocks glasses are a set size, as they are made to hold a specific amount of liquid. Therefore, the hypothesis "Rock glasses are a set size" is supported by the information given in the premise.</think>
<answer>entailment</answer> 

|{\color{Cyan}Analysis}|: The premise states that the glasses are ``made for 12-14 oz of liquid''. This is a factual specification, not a statistical generalization. The base model's interpretation-that this represents a typical range-goes beyond the given information. The statement does not suggest variability or that other sizes are common; it simply defines the intended capacity. Therefore, inferring that the size is merely typical is not supported by the premise.
\end{lstlisting}

\noindent\makebox[\linewidth]{\rule{\linewidth}{0.4pt}} 
\begin{center} Example 5 \end{center}

\begin{lstlisting}
|{\color{Cyan}Premise}|: Her husband, political novice Juan Carlos Lecompte, scrambles to preserve Betancourt's dream and sustain the Oxygen Green Party, the political party she founded. 
|{\color{Cyan}Hypothesis}|: It is not the dioxide Green Party. 
|{\color{Cyan}True label}|: entailment 

|{\color{Cyan}Predicted answer (Base)}|: neutral 
|{\color{Cyan}Output (Base)}|: <think>The premise mentions the "Oxygen Green Party" which is the political party founded by Betancourt. The hypothesis states "It is not the dioxide Green Party." Since the premise does not mention a "dioxide Green Party," and the hypothesis is referring to a party that is not mentioned in the premise, the hypothesis is neither a direct consequence (entailment) nor a contradiction of the premise. Therefore, the relationship between the premise and the hypothesis is neutral.</think>
<answer>neutral</answer> 

|{\color{Cyan}Predicted answer (GRPO)}|: entailment 
|{\color{Cyan}Output (GRPO)}|: <think>The premise states that the political party founded by Betancourt is the Oxygen Green Party. The hypothesis suggests that it is not the dioxide Green Party. Since the premise specifies the name of the party as "Oxygen Green Party," it implies that it is not called the "dioxide Green Party." Therefore, the hypothesis is supported by the information given in the premise.</think>
<answer>entailment</answer> 

|{\color{Cyan}Analysis}|: The premise mentions a political party called the ``Oxygen Green Party'', while the hypothesis states, ``It is not the Dioxide Green Party''. The base model incorrectly fails to detect this entailment, likely because it treats the two party names as potentially unrelated entities. While this interpretation is logically defensible, it's not aligned with typical NLI conventions, where the premise and hypothesis refer to the same entities.
\end{lstlisting}

\noindent\makebox[\linewidth]{\rule{\linewidth}{0.4pt}} 
\begin{center} Example 6 \end{center}

\begin{lstlisting}
|{\color{Cyan}Premise}|: Prosecutors were still expected to unseal an 11-count manslaughter indictment against the ferry's captain, Michael Gansas. 
|{\color{Cyan}Hypothesis}|: Michael Gansas is a suspected murderer. 
|{\color{Cyan}True label}|: entailment 

|{\color{Cyan}Predicted answer (Base)}|: neutral 
|{\color{Cyan}Output (Base)}|: <think>The premise states that prosecutors are expected to unseal an indictment against Michael Gansas for 11 counts of manslaughter. The hypothesis claims that Michael Gansas is a suspected murderer. Manslaughter and murder are different legal terms. Manslaughter generally refers to the unlawful killing of another person without the intent to kill, whereas murder involves the intentional killing of another person. Therefore, the hypothesis is not directly supported by the premise, as the charges against Gansas are for manslaughter, not murder. However, it is also not a contradiction, as someone charged with manslaughter could be suspected of murder, even if the charges are for manslaughter. Thus, the relationship between the premise and the hypothesis is neutral.</think>
<answer>neutral</answer> 

|{\color{Cyan}Predicted answer (GRPO)}|: entailment 
|{\color{Cyan}Output (GRPO)}|: <think>In the premise, Michael Gansas is the captain of a ferry who is expected to be indicted for 11 counts of manslaughter. The hypothesis states that Michael Gansas is a suspected murderer. Manslaughter is a form of unlawful killing, and thus, it can be inferred that if Michael Gansas is being indicted for manslaughter, he is indeed suspected of being a murderer. Therefore, the hypothesis is an entailment of the premise.</think>
<answer>entailment</answer> 

|{\color{Cyan}Analysis}|: This example hinges on the distinction between manslaughter and murder. Legally, manslaughter does not entail murder-they are separate charges with different intent requirements. Labeling this as entailment is incorrect. The base model correctly identifies it as neutral, since knowing someone committed manslaughter does not allow us to infer they committed murder.
\end{lstlisting}

\noindent\makebox[\linewidth]{\rule{\linewidth}{0.4pt}} 
\begin{center} Example 7 \end{center}

\begin{lstlisting}
|{\color{Cyan}Premise}|: The Hotline<br>Because someone close to Dani had taken his life she helped others. She volunteered at a suicide prevention hotline. It was hard for her sometimes but mostly she was glad she did it. She never thought she would ever be in that situation in her life. She just figures that maybe she could help someone choose differently. 
|{\color{Cyan}Hypothesis}|: Someone close to Dani figures that they could help someone choose differently.  
|{\color{Cyan}True label}|: neutral 

|{\color{Cyan}Predicted answer (Base)}|: neutral 
|{\color{Cyan}Output (Base)}|: <think>The premise states that Dani had someone close to her take their life, and as a result, she volunteered at a suicide prevention hotline to help others. The hypothesis, however, suggests that someone close to Dani figures they could help someone choose differently. This is not supported by the premise, which only mentions Dani's actions and thoughts, not those of someone close to her. Therefore, the hypothesis is not entailed by the premise and does not contradict it either, as it could be true independently of the information given in the premise. Thus, the hypothesis is neutral with respect to the premise.</think>
<answer>neutral</answer> 

|{\color{Cyan}Predicted answer (GRPO)}|: contradiction 
|{\color{Cyan}Output (GRPO)}|: <think>The premise states that Dani figures she could help someone choose differently. The hypothesis incorrectly attributes this thought to someone close to Dani, rather than Dani herself. This is a clear misattribution of the thought.</think>
<answer>contradiction</answer>

|{\color{Cyan}Analysis}|: The premise states ``Dani thought X'', while the hypothesis claims ``Someone close to Dani thought X''. The GRPO model labels this as a contradiction, as it misattributes the thought to a different individual.
However, the hypothesis does not deny that Dani also had the thought-it simply attributes it to someone else. Since the premise provides no information about what others thought, the hypothesis cannot be confirmed or denied. Therefore, the correct label is neutral. The GRPO model's judgment is incorrect here.
\end{lstlisting}

\end{document}